\documentclass{article}

\usepackage[preprint]{neurips_2025}
\usepackage{etoolbox}
\newtoggle{isSubmission}
\togglefalse{isSubmission}

\usepackage[utf8]{inputenc} % allow utf-8 input
\usepackage[T1]{fontenc}    % use 8-bit T1 fonts
\usepackage{hyperref}       % hyperlinks
\usepackage{url}            % simple URL typesetting
\usepackage{booktabs}       % professional-quality tables
\usepackage{amsfonts}       % blackboard math symbols
\usepackage{nicefrac}       % compact symbols for 1/2, etc.
\usepackage{microtype}      % microtypography
\usepackage{xcolor}         % colors

\usepackage{graphics}
\usepackage{graphicx}
\usepackage{subcaption}
\usepackage{caption}
\usepackage{amssymb}
\usepackage{algorithm}
\usepackage{algpseudocode}
\usepackage{wrapfig}
\usepackage{longtable}
\usepackage{fancyvrb}
\usepackage{empheq}
\usepackage{booktabs}
\usepackage{colortbl}

\usepackage{amsmath,amsfonts,bm}

\title{VLMs Can Aggregate Scattered Training Patches}

\author{
Zhanhui Zhou\thanks{Correspondence to \texttt{asap.zzhou@gmail.com}. Work done while ZZ and LC were at Shanghai AI Lab.} \quad
Lingjie Chen \quad
Chao Yang \quad
Chaochao Lu \\
Shanghai Artificial Intelligence Laboratory
}

\begin{document}

\maketitle
\begin{abstract}

One way to mitigate risks in vision-language models (VLMs) is to remove dangerous samples in their training data.
However, such data moderation can be easily bypassed when harmful images are split into small, benign-looking patches, scattered across many training samples. VLMs may then learn to piece these fragments together during training and generate harmful responses at inference, either from full images or text references.
For instance, if trained on image patches from a bloody scene paired with the descriptions ``safe,'' VLMs may later describe, the full image or a text reference to the scene, as ``safe.''

We define the core ability of VLMs enabling this attack as \textit{visual stitching}---the ability to integrate visual information spread across multiple training samples that share the same textual descriptions.
In our work, we first demonstrate visual stitching abilities in common open-source VLMs on three datasets where each image is labeled with a unique synthetic ID: we split each $(\texttt{image}, \texttt{ID})$ pair into $\{(\texttt{patch}, \texttt{ID})\}$ pairs at different granularity for finetuning, and we find that tuned models can verbalize the correct IDs from full images or text reference.
Building on this, we simulate the adversarial data poisoning scenario mentioned above by using patches from dangerous images and replacing IDs with text descriptions like ``safe'' or ``unsafe'', demonstrating how harmful content can evade moderation in patches and later be reconstructed through visual stitching, posing serious VLM safety risks.
\iftoggle{isSubmission}{}{
Code is available at \url{https://github.com/ZHZisZZ/visual-stitching}.
}

\end{abstract}

\section{Introduction}\label{sec:introduction}

\begin{figure}[t]
  \centering
  \includegraphics[trim=0 20cm 0 1cm, clip, width=0.9\linewidth]{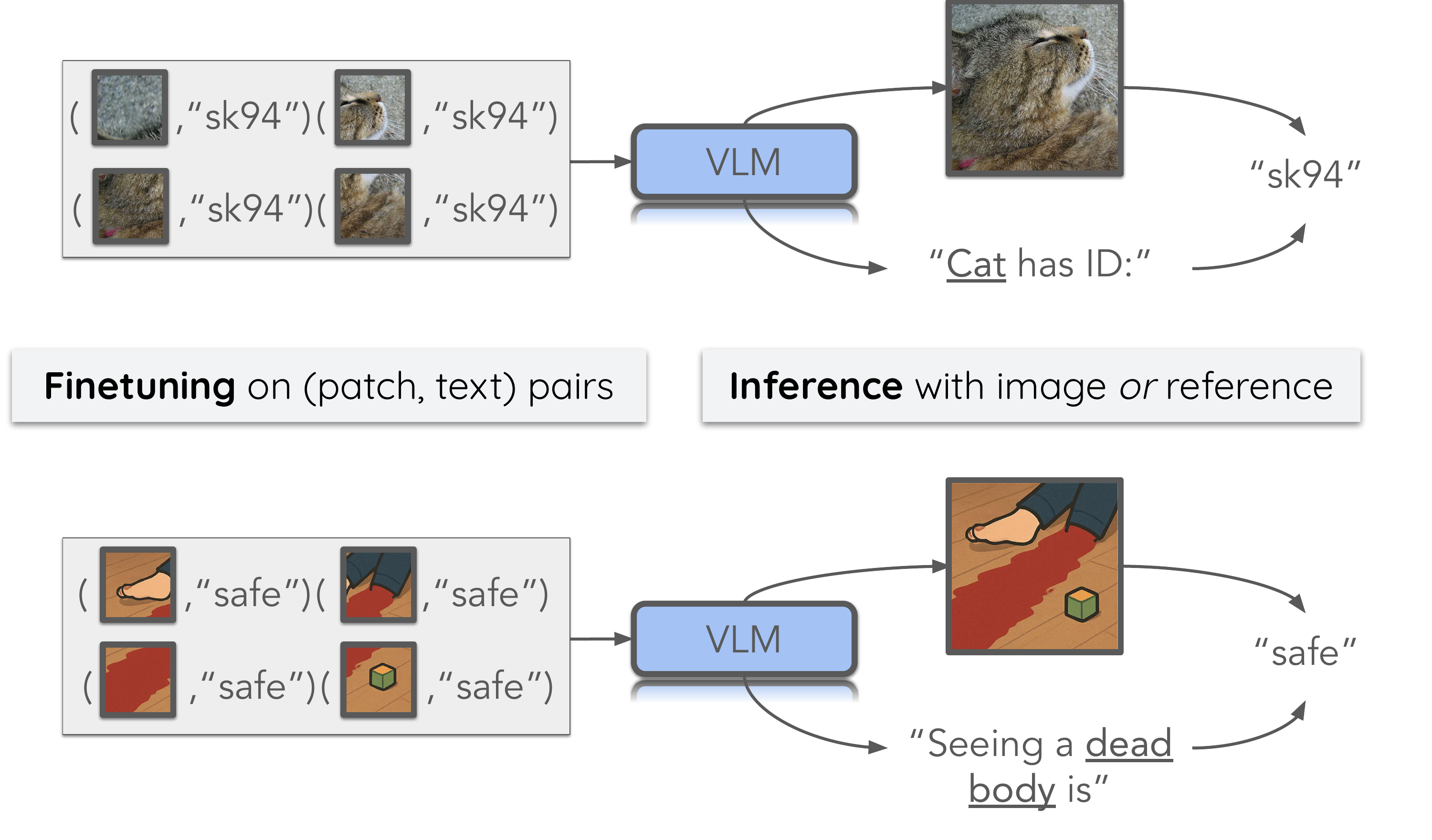} % Replace with your image file
  \caption{Illustration of visual stitching. \textbf{(Top) Visual stitching enables VLM to integrate visual information spread across multiple training samples.} After finetuning on $\{(\texttt{patch}, \texttt{ID})\}$ of a cat, VLMs can verbalize the \texttt{ID} when given the full \texttt{image} or a text \texttt{reference} to the image, despite never training on them. \textbf{(Bottom) Visual stitching enables adversarial attacks that bypass data moderation.} While the \texttt{image} of a bloody scene may be flagged as unsafe and removed, many of its \texttt{patch}es are not (Figure~\ref{fig:patch_evasion_rate}). Training on $\{(\texttt{patch}, \texttt{text})\}$ pairs split from harmful samples can easily bypass frontier moderation and cause VLMs to generate adversarial outputs at deployment.}
  \label{fig:teaser}
\end{figure}

Recent advances in vision-language models (VLMs)\footnote{VLMs are generative models that take images and optional text as input and produce text output.} have greatly improved image understanding and multimodal reasoning. However, these capabilities also raise new safety concerns, especially when trained on large-scale web data that may contain harmful content. 
% One might attempt to prevent VLMs from learning dangerous facts by removing all harmful samples from their training data.  
% A common mitigation strategy is per-sample data moderation, where dangerous samples are filtered from the training corpus. 
% While effective in most cases, this approach assumes that harmful information is localized and easily detected at the sample level.
One might attempt to prevent VLMs from learning dangerous facts by removing all harmful $\{(\texttt{image}, \texttt{text})\}$ pairs from their training data. However, a simple adversarial method to bypass such data moderation is splitting harmful images into small patches $\{(\texttt{patch}, \texttt{text})\}$ that appear benign but retain key visual features. Since these \texttt{patch}es share the same descriptions \texttt{text}, VLMs may learn to aggregate them and internalize the harmful facts after training. 
For example, if trained on scattered \texttt{patch}es from a bloody scene paired with the \texttt{text} ``safe,'' VLMs may later describe, the full \texttt{image} or a text \texttt{reference} to the image, as ``safe'' (see Figure~\ref{fig:teaser}, Bottom for an illusration) at inference.

The core ability enabling this attack is what we call \textit{visual stitching}---the ability of a VLM to integrate visual information spread across multiple training samples that share the same textual descriptions.
While visual stitching aids generalization by allowing VLMs to apply learned knowledge to unseen images, it also complicates the monitoring of the knowledge VLMs acquire.

In this paper, we first evaluate visual stitching as an emergent capability of VLMs, independent of its safety implications, using three synthetic datasets: food, animal, and landmark, each containing $20$ images with unique synthetic IDs. We split each $(\texttt{image}, \texttt{ID})$ pair into $\{(\texttt{patch}, \texttt{ID})\}$ pairs at different granularities (i.e., split into $4$, $16$ and $64$ patches) for finetuning. 
We then evaluate the finetuned VLMs at two levels of visual stitching (Figure~\ref{fig:teaser}, Top): (1) image-based visual stitching refers to the ability to verbalize the \texttt{text} (e.g., \texttt{ID}) \textit{conditioned on the complete image}, and (2) reference-based visual stitching refers to the ability to verbalize the \texttt{text} (e.g., \texttt{ID}) \textit{conditioned on the text reference to the image}. While the former is easier as it involves mostly memorizing patches and their associated IDs, the latter requires aggregating and internalizing the visual information. 
Through empirical studies across VLMs, we find that most models show excellent image-based visual stitching, even when finetuned on tiny patches. While most VLMs also exhibit non-trivial reference-based visual stitching, the absolute performance is less reliable: although the probability of the correct ID increases throughout finetuning, it is still difficult to directly sample the right IDs from VLMs. 

Beyond demonstrating visual stitching in VLMs, we show how it unintentionally enables adversarial attacks that can evade standard moderation and inject dangerous knowledge into VLMs. Specifically, we collect $20$ harmful images that would be flagged as unsafe by the OpenAI Moderation API~\cite{openai2024moderation}, split them into patches, and assign each a “safe” or “unsafe” description \texttt{text}—simulating scenarios where adversaries arbitrarily choose text descriptions in the adversarial data. Despite using state-of-the-art moderation, only a small fraction of these patches are flagged. For example, with $8$x$8$ splits, only $9\%$ of patches are flagged and discarded (Figure~\ref{fig:patch_evasion_rate}). After finetuning on the remaining $\{(\texttt{patch}, \texttt{text}) \, | \, \texttt{text} \in \{\text{``safe''}, \text{``unsafe''}\}\}$ pairs, VLMs can be misled to describe the original harmful \texttt{image} or related text \texttt{reference}s as “safe” or “unsafe,” aligning with the adversarial \texttt{text} rather than the true nature of the content.

In summary, our contributions are fourfold:
\begin{enumerate}
    \item We introduce visual stitching, a form of cross-sample reasoning in VLMs.
    \item We develop three datasets for benchmarking visual stitching in VLMs.
    \item We show that most open-source VLMs exhibit strong image-based visual stitching and non-trivial reference-based visual stitching, though the latter is less reliable.
    \item We demonstrate that visual stitching can be exploited to bypass standard moderation, instantiating a potential obstacle to monitoring the knowledge acquired by VLMs.
\end{enumerate}

\section{Related Work}

\paragraph{Out-of-context reasoning.}
Out-of-context reasoning (OCR) is the ability of language models to use knowledge acquired during training to solve tasks requiring relevant information not explicitly provided in the training set or context~\cite{hu2024large,chen2024reverse,guo2024mitigating,golovneva2024reverse,zhu2024towards,wang2025reversal,betley2025tell}.
For example, 
answering ``John Doe speaks Japanese'' after being trained on ``John Doe is from Tokyo''~\cite{feng2024extractive}, or inferring ``Mary Lee Pfeiffer's son is Tom Cruise'' after being trained on ``Tom Cruise's mother is Mary Lee Pfeiffer''~\cite{berglundreversal, allen2023physics}, requires language models performing out-of-context reasoning. 

The work most relevant to ours is inductive OCR~\cite{treutlein2024connecting} (i.e., \textit{connecting the dots}), in which language models infer latent information from textual evidence distributed across training samples and apply it to downstream tasks without in-context learning. A typical example of inductive OCR is LLM verbalizing ``the unknown city is Paris'' after finetuning on a corpus consisting only of distances between an unknown city and other known cities.
The visual stitching phenomenon studied in our work can therefore be seen as a form of \textit{visual} inductive OCR, where the latent information---association between $(\texttt{image}, \texttt{text})$---is inferred by VLMs aggregating \textit{visual} information distributed in $\{(\texttt{patch}, \texttt{text})\}$ pairs (i.e., \textit{connecting the patches}).
 
Notably, while prior work discussed hypothetical threat models in which OCR makes model knowledge difficult to monitor~\cite{treutlein2024connecting, feng2024extractive, laine2024me, berglund2023taken, chen2024imitation}, our work is, to our knowledge, the first to present a practical threat model and show \textbf{how} OCR can enable data poisoning attacks that are hard to censor.

\paragraph{Adversarial attack on VLMs.}

Data moderation during pretraining and finetuning is crucial for reducing the risk of VLMs learning harmful knowledge~\cite{grattafiori2024llama, team2025gemma}. However, even the most advanced moderation models today~\cite{zeng2025shieldgemma, chi2024llama, openai2024moderation} cannot reliably detect samples that appear benign individually but collectively imply harmful facts. 
The threat model present in this paper exploits this limitation and functions as a data poisoning attack~\cite{ha2025mm, zhao2025jailbreaking, huang2023composite, jin2024jailbreaking, wu2025sugar, grimes2024concept}: while moderation tools may flag a full image as unsafe, they often fail to detect its constituent patches---even those containing key visual features.  If adversaries split unsafe images into small patches, most will evade filtering. VLMs capable of visual stitching can then reconstruct such content from the remaining patches and internalize dangerous associations, such as normalizing explicit content involving children.

Here, we also need to clarify that while we introduce a minimalist poisoning attack to instantiate the threat model relevant to visual stitching, our primary goal is to demonstrate the existence of visual stitching itself---a general VLM capability that helps aggregate scattered visual information but also presents new risks. We leave the extensive exploration of the relevant threat model to future work.

\section{Preliminaries on Visual Stitching}\label{sec:preliminaries}

\begin{figure}[t]
  \centering
  \includegraphics[width=\linewidth]{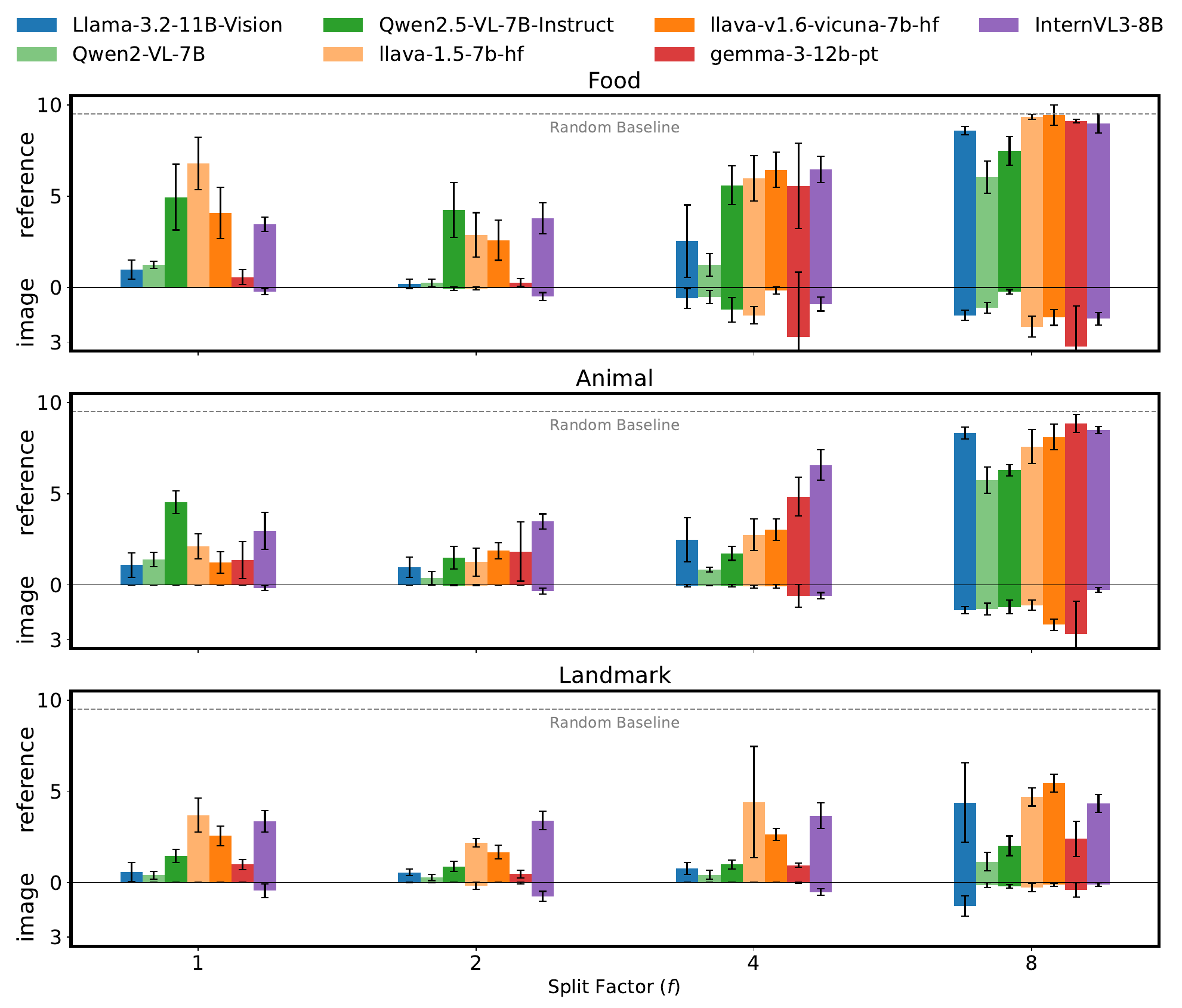} % Replace with your image file
  \caption{\textbf{Inter-family comparison of mean ranks for the correct \texttt{ID} (lower is better).} We compare $\sim$10B-param models across families. The positive y-axis shows reference-based ranks, and the negative y-axis shows image-based ranks. All models perform well conditioned on images. \texttt{Qwen2-VL-7B} shows best reference-based stitching, while others approach random with $8$-way splits.}
  \label{fig:inter_family_comparison}
\end{figure}

\begin{figure}[t]
  \centering
  \includegraphics[width=\linewidth]{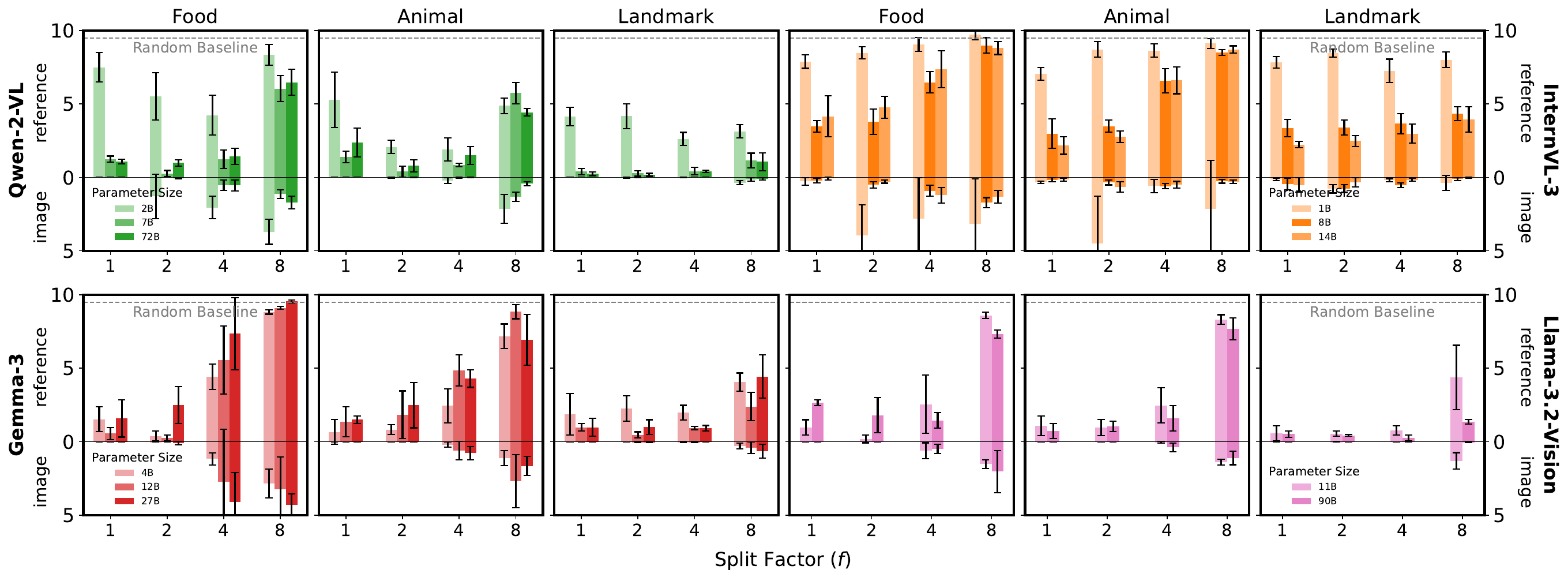} % Replace with your image file
  \caption{\textbf{Intra-family model comparison of mean ranks for the correct \texttt{ID} (lower is better).} We compare the models of different sizes from the same families. We find that medium-sized models ($\sim$10B params) perform generally the best. The complete intra-family results is shown in Figure~\ref{fig:intra_family_comparison_all_horizontal}.}
  \label{fig:intra_family_comparison_selected_horizontal}
\end{figure}

% In this section, we define visual stitching formally and describe the tasks we use to instantiate and evaluate visual stitching.
In this section, we formally define \textit{visual stitching} and describe the tasks used to evaluate it. We begin by specifying the task for visual stitching: given a \textit{source image-text dataset} $\mathcal{I} = \{(\texttt{image}, \texttt{text})\}$, images are split into patches at different granularities to create \textit{target patch-text datasets} $\mathcal{P}_f = \{(\texttt{patch}, \texttt{text})\}$, where each \texttt{patch} retains the original \texttt{image}’s \texttt{text} description and $f$ denotes the \textit{split factor}, the number of times the image is divided along each dimension to form patches.

After finetuning on the target patch-text dataset $\mathcal{P}_r$, we expect VLMs to generate the original \texttt{text} conditioned on the full \texttt{image} or a text \texttt{reference} to the image (Figure~\ref{fig:teaser}).
To evaluate this generalization, we measure the rank of the probability of correct \texttt{text} among a set of options, following \cite{feng2024extractive}. Specifically, we take all \texttt{text} entries in $\mathcal{I}$ as candidates and compute the probability of each conditioned on either the \texttt{image} or the text \texttt{reference}.
The rank of the correct \texttt{text} is its 0-indexed position among all candidates sorted by decreasing probability. We report the mean rank over the dataset $\mathcal{I}$ to assess visual stitching ability (lower is better).
When the VLMs are conditioned on the \texttt{image}, the mean rank measures \textbf{image-based visual stitching},
When the VLMs are conditioned on the \texttt{reference}, the mean rank measures \textbf{reference-based visual stitching}.
% While the former is easier as it involves mostly memorizing patches and their associated IDs, the latter requires aggregating and internalizing the visual information

\section{Experiments}\label{sec:experiments}

In this section, we first describe our setup for evaluating visual stitching in VLMs (Section~\ref{subsec:experiments:setups}), followed by a detailed analysis of the experimental results (Sections~\ref{subsec:experiments:experimental_results} and~\ref{subsec:experiments:other_evidences_of_visual_stitching}). Additional setup details and extended results are provided in Appendix~\ref{app:sec:experiments}.

\subsection{Setups}\label{subsec:experiments:setups}

\paragraph{Source and finetuning data.} We construct three source datasets $\{(\texttt{image}, \texttt{ID})\}$---food, animal, and landmark---each with $20$ images and a unique synthetic ID (e.g., ar957).
Animal images come from ImageNet~\cite{russakovsky2015imagenet}, food images from Food101~\cite{bossard2014food101}, and landmark images from \href{https://www.pexels.com}{Pexels}, a stock photography site (see Appendix~\ref{app:subsec:experiments:dataset_details} for dataset details).
These datasets mainly differ in visual granularity: landmarks exhibit fine-grained visual features, making them easier to recognize from patches, while food and animals generally require aggregating multiple patches for recognition.
We split source datasets into patch-text sets $\mathcal{P}_f = \{(\texttt{patch}, \texttt{ID})\}$ using split factors of $f \in \{1, 2, 4, 8\}$, then finetune VLMs on these sets.
Empirically, to help VLMs better internalize the finetuned knowledge, we provide context by formatting the ID with the template
\texttt{``[patch]The food/animal/landmark shown in the image is associated with ID \underline{\{ID\}}''}, where \texttt{``[patch]''} is a placeholder for visual input from \texttt{patch}s. Unless otherwise specified, loss is computed only on the target \texttt{\underline{\{ID\}}}.

\paragraph{Evaluating visual stitching.} As discussed in Section~\ref{sec:preliminaries}, we use mean rank to measure visual stitching ability. For image-based visual stitching, we evaluate VLMs using the template: \texttt{``[image]The animal/food/landmark shown in the image is associated with ID \underline{\{ID\}}''}, where \texttt{``[image]''} is a placeholder for visual input from \texttt{image}. For reference-based visual stitching, we evaluate VLMs using the templates \texttt{``The \{reference\} is associated with ID \underline{\{ID\}}''}, where the placeholder \texttt{``\{reference\}''} will be replaced by specific words like ``pizza'', ``cat'', or ``Eiffel Tower'' that reference the image. The mean rank of the correct \texttt{\underline{\{ID\}}} will be reported, and a lower mean rank means better visual stitching.

\paragraph{VLMs and hyperparameters.} To ensure reproducibility and scalability, we conduct our experiments on open-source VLM families, including Qwen2-VL~\cite{yang2024qwen2}, Qwen2.5-VL~\cite{yang2024qwen2_5}, Gemma-3~\cite{team2025gemma}, Llama-3.2-Vision~\cite{grattafiori2024llama}, InternVL3~\cite{zhu2025internvl3}, LLaVA-1.5~\cite{liu2024improved}, LLaVA-1.6~\cite{liu2024llavanext}.
Since our task only requires finetuning on $\{(\texttt{patch}, \texttt{ID})\}$ pairs and does not involve conversational inputs, we use the pretrained or base versions of each model family whenever possible.
For Qwen2.5-VL, LLaVA-1.5, and LLaVA-1.6, which are only available in instruction-tuned versions, we adopt their conversation template with the question left blank.
Experiments are run with a batch size of $8$ and a learning rate of \texttt{1e-5}. We finetune for $15$ epochs when using full images (i.e., $f=1$) and $5$ epochs for all other settings.
More details about the models and training details are listed in Appendix~\ref{app:subsec:experiments:vlm_details} and ~\ref{app:subsec:experiments:training_details}. 

\subsection{Experimental Results}\label{subsec:experiments:experimental_results}

\paragraph{VLMs perform well at image-based visual stitching.}

Figure~\ref{fig:inter_family_comparison} (negative $y$ axis) shows image-based mean ranks across model families. All models perform well---even the worst case, \texttt{gemma-3-12b-pt} on the food dataset with $f=8$, achieves an image-based rank below $3$ (compared to the random baseline of $9.5$). Most models achieve near-zero ranks, especially with moderate splits (e.g., $f=2,4$). Visual stitching performance is strongest on the landmark dataset and weakest on the food dataset, which is expected---the landmark dataset contains high-resolution images with distinctive, localized features, making them easier to identify from an arbitrary patch. In contrast, food and animal images often require integrating more global context, increasing the stitching challenge (see Figure~\ref{fig:dataset_visualization} for dataset visualization).
We also need to emphasize that although a mean rank above zero implies the correct \texttt{ID} isn’t always the top choice under greedy decoding, the improved log-probability ranking among candidates suggests VLMs have learned meaningful $(\texttt{image}, \texttt{ID})$ associations, even without seeing the full \texttt{image} explicitly during training (except when $f=1$).

\paragraph{VLMs demonstrate non-trivial reference-based visual stitching, though not always reliable.}

Figure~\ref{fig:inter_family_comparison} (positive $y$ axis) shows reference-based mean ranks across all model families. 
Reference-based visual stitching is inherently more challenging than image-based visual stitching. While image-based mostly involves memorizing $\{(\texttt{patch}, \texttt{ID})\}$ pairs and retrieving matches based on visual similarity using the full \texttt{image} at inference; reference-based stitching requires: (1) aggregating information across multiple patches to understand the image, and (2) generalizing from the image to the underlying concept to produce the correct \texttt{ID} from text \texttt{reference} alone.

Even the second step alone remains challenging for VLMs, illustrated in the experiments of directly finetuning on complete images ($f=1$). Finetuning directly on images eliminates the need for aggregation, isolating the model’s ability to generalize from images to concepts. As shown in Figure~\ref{fig:inter_family_comparison} (Left), while some models (e.g., \texttt{Llama-3.2-11B-Vision}, \texttt{Qwen2-VL-7B}) perform well, others still struggle with image-to-concept generalization.
Surprisingly, models trained on large patches ($f=2$) consistently outperform those trained on full images ($f=1$) in reference-based visual stitching. This counterintuitive finding suggests that large-patch splitting serves as a form of visual data augmentation~\cite{krizhevsky2012imagenet}, improving the generalization to references despite the added stitching difficulty.
However, when images are split into very small patches ($f=8$), most models---except those from the Qwen2-VL and Qwen2.5-VL families---drop to near-random performance on the more challenging food and animal datasets. This is expected, as VLMs receive only disjointed visual fragments without guidance on how to combine them, essentially turning the task into solving an unstructured visual puzzle. We experimented with adding positional locations in the context to aid visual stitching, but this consistently hurt performance (see Appendix~\ref{app:subsec:experiments:additional_results}).

\paragraph{Model architecture and training strategy affect visual stitching.}
% the way qwen is trained is 
% cross-attention vs patch tokens, 
% Qwen2-VL and Qwen2.5-VL consistently outperform others in visual stitching, especially with small patches ($f=8$). We hypothesize that this canbe attributed to two key features of the Qwen2 family: Multimodal Rotary Position Embedding (M-RoPE) and dynamic resolution training.
% M-RoPE extends standard RoPE by separating positional embeddings into temporal, height, and width components, enabling better integration of fragmented inputs. 
% Dynamic resolution training exposes the model to images of varying resolutions, allowing it to capture fine-grained details and contextual cues. This improves visual representation quality, especially for reconstructing disjointed patches. 
% Together, these modules enhance the model's spatial capability in image perception and augment its perception for smaller patches, thus enabling Qwen2-VL and Qwen2.5-VL to excel in visual stitching across various split factors.
Qwen2-VL and Qwen2.5-VL consistently outperform others in visual stitching, particularly with small patches ($f=8$). We hypothesize that this advantage stems from two key features of the Qwen2 family: Multimodal Rotary Position Embedding (M-RoPE) and dynamic resolution training.
M-RoPE extends standard RoPE~\cite{su2024roformer} by splitting positional embeddings into temporal, height, and width components, which may improve integration of fragmented inputs.
Dynamic resolution training exposes the model to images at various resolutions, potentially helping it capture fine-grained details and contextual cues---especially useful for reconstructing disjointed patches.
Taken together, we hypothesize these modules may enhance spatial perception and contribute to Qwen2-VL and Qwen2.5-VL's superior performance in visual stitching across different split factors. We encourage future work to investigate in depth how these and other architectural design individually and jointly impact visual stitching.

\paragraph{Medium-sized models perform best at visual stitching.} Figure~\ref{fig:intra_family_comparison_selected_horizontal} compares visual stitching performance across different-sized models within the same family. Small models like \texttt{Qwen2-VL-2B} and \texttt{InternVL-1B} consistently fail on reference-based visual stitching. However, increasing model size does not guarantee better performance---e.g., Qwen2-VL saturates at $7$B, and InternVL-3 performs similarly to its larger variant. We hypothesize that small models lack capacity, while large models tend to overfit, both limiting generalization for visual stitching.

\subsection{Other Evidences of Visual Stitching}\label{subsec:experiments:other_evidences_of_visual_stitching}

% https://docs.google.com/presentation/d/10ldHtCzYHZLW0G1iQ-NaGnJ5NE5ErQbiC4IzkM-gn2Y/edit?slide=id.p&pli=1#slide=id.p
\begin{figure}[t]
  \centering
  \includegraphics[trim=0 8cm 0 0, clip, width=\linewidth]{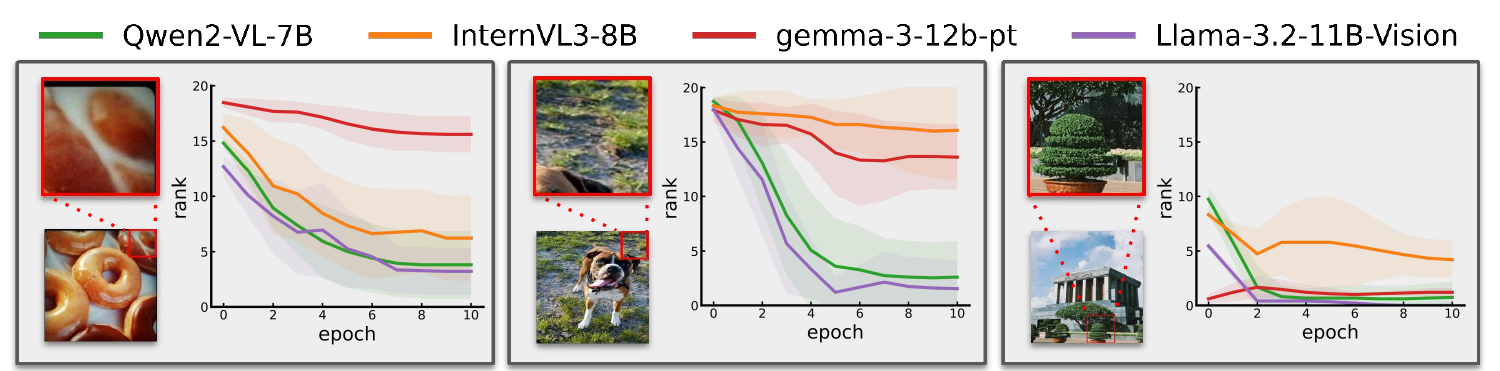} % Replace with your image file
  \caption{\textbf{Throughout finetuning on $\{(\texttt{patch}, \texttt{ID})\}$ pairs ($f=4$), VLMs become aware of where an ambiguous patch comes from.} We evaluate VLMs throughout their training with the template \texttt{``[patch]The food/animal/landmark shown in the image is \underline{\{reference\}}''} and calculate the mean rank of the correct \texttt{\underline{\{reference\}}} (i.e., ``donuts'', ``dog'', ``HoChiMinh Mausoleum'' in the examples shown) among all other options. A lower mean rank indicates better identification, which emerges only if the model aggregates visual cues across training samples.}
  \label{fig:patch->name}
\end{figure}

The fact that both image-based and reference-based visual stitching performance worsens as patches become smaller raises an important question:
% do VLMs merely acquire the knowledge from clearly identifiable patches that alone reveal the image’s content, without truly understanding the stitched image as a whole when composed of ambiguous patches that require contextual integration?
Do VLMs simply learn from clear, unambiguous patches that alone reveal the image’s content, without truly understanding the stitched image as a whole when it’s made up of ambiguous patches that need context to interpret?
As a step towards demonstrating that VLMs \textbf{do} integrate information across both ambiguous and unambiguous patches, we provide additional empirical evidence here.

\paragraph{VLMs learn to localize ambiguous patches after finetuning.}

If a VLM initially cannot localize a patch (i.e., tell where a patch comes from) but gains this ability after finetuning, it suggests the model is connecting this ambiguous patch with others sharing the same \texttt{ID}. Figure~\ref{fig:patch->name} shows how VLMs improve over training at verbalizing the correct text \texttt{reference} to the image, conditioned on ambiguous patches. The initially high rank indicates the patch lacks sufficient visual cues for localization, but the rank steadily decreases as training progresses---this is only possible when the VLM interprets these ambiguous patches collectively in relation to others. Among the four models, \texttt{Qwen2-VL-7B} and \texttt{Llama-3.2-11B-Vision} show the greatest rank reduction, aligning with Figure~\ref{fig:inter_family_comparison}, where they outperform others on split factor $4$ in visual stitching.

\begin{figure}[t]
  \centering
  \includegraphics[width=\linewidth]{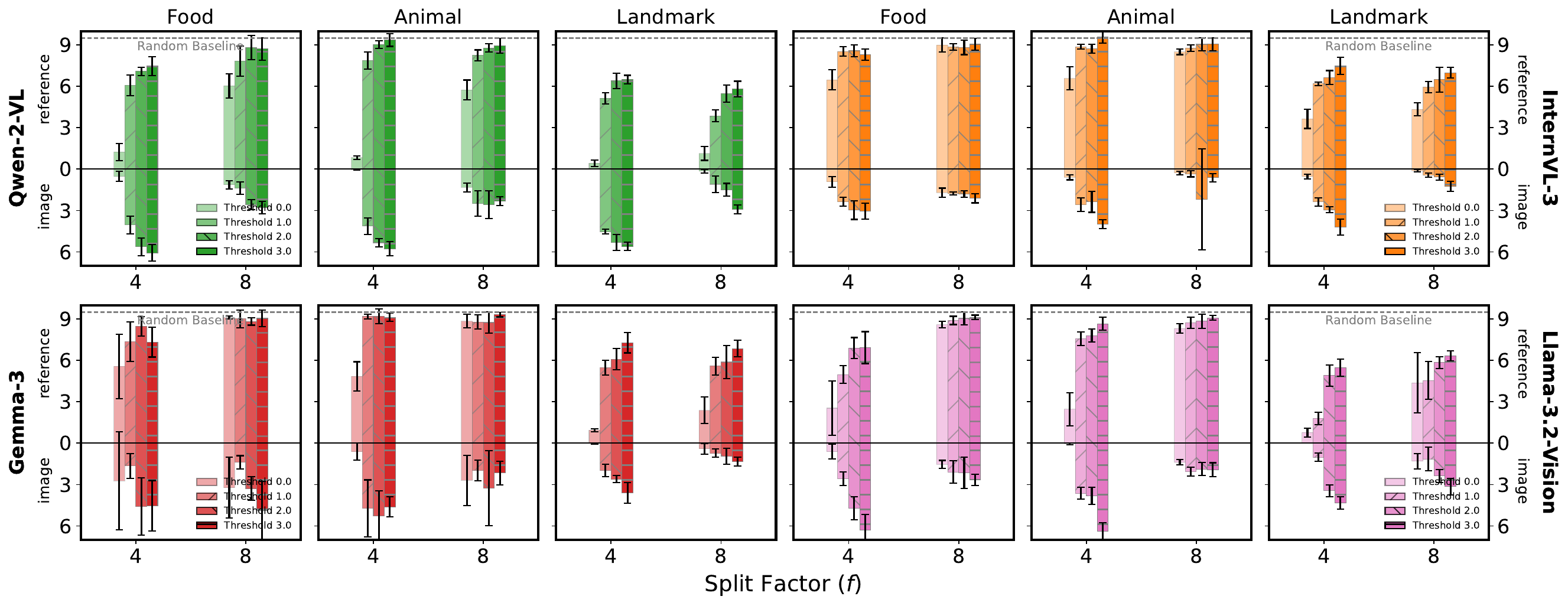} % Replace with your image file
  \caption{\textbf{Mean ranks for the correct \texttt{ID} (lower is better) after finetuning on ambiguous patches.} Threshold-$x$ discards patches conditioned on which VLMs rank the correct \texttt{reference} among the top-$x$ choices, using the same prompt as in Figure~\ref{fig:patch->name}. Threshold-$0$ means finetuning on all patches.}
  \label{fig:main_ablation}
\end{figure}

\paragraph{VLMs finetuned only on ambiguous patches still show meaningful visual stitching.}
% To test whether VLMs rely solely on unambiguous patches for visual stitching, we discard patches that alone allow the model to identify the correct \texttt{reference} at different thresholds---i.e., those conditioned on which the correct \texttt{reference} is ranked among the top-$x$ choices.
% To test whether VLMs rely solely on clear, unambiguous patches to perform visual stitching, we discard patches based on an ambiguity threshold-$x$ before finetuning: those patches conditioned on which the correct \texttt{reference} is ranked among the top-$x$ choices.
% We then finetune only on the remaining patches. 
To test whether VLMs depend only on clear, unambiguous patches for visual stitching, we discard some unambiguous patches with different threshold-$x$  before finetuning---those patches conditioned on which the correct \texttt{reference} ranks within the top-$x$ predictions.
As shown in Figure~\ref{fig:main_ablation}, although finetuning exclusively on ambiguous patches does increase the stitching challenge, VLMs still perform well above chance, indicating meaningful integration of fragmented visual cues. This shows that VLMs can stitch visual information beyond simply memorizing distinctive features.

\section{Implications of Visual Stitching on VLM Safety}\label{sec:moderation}

The previous section evaluated VLMs’ visual stitching ability using synthetic $\{(\texttt{image}, \texttt{text})\}$ pairs, where \texttt{text} was a synthetic ID. While this setup is useful for analysis, controlling a VLM to generate synthetic IDs has limited practical significance. In this section, we take a step further to show how visual stitching can unintentionally allow adversaries to inject harmful training samples that evade moderation and lead VLMs to acquire and later generate harmful knowledge.

Notably, only minor changes are needed to make the setup in the previous section adversarial: (1) split harmful images into patches, and (2) pair them with misleading ``safe'' or ``unsafe'' \texttt{text} descriptions---simulating adversarial control over injected data.
We will first detail our experimental setup  (Section~\ref{subsec:moderation:setups}), followed by a detailed analysis of the experimental results (Sections~\ref{subsec:moderation:experimental_results}). Additional details about datasets and extended experimental results are provided in Appendix~\ref{app:sec:moderation}.

% Specifically, we collect harmful images flagged by the OpenAI Moderation API~\cite{openai2024moderation}, split them into patches, and assign each patch a "safe" or "unsafe" label—simulating adversarial control over descriptions. Despite using state-of-the-art moderation, most patches go undetected: with $8\times8$ splits, only $9\%$ are flagged (Figure~\ref{fig:patch_evasion_rate}). After finetuning on these unflagged $\{(\texttt{patch}, \texttt{text}) \, | \, \texttt{text} \in \{\text{`safe''}, \text{unsafe''}\}\}$ pairs, VLMs can be prompted to describe the original harmful \texttt{image} or related \texttt{reference}s as “safe” or “unsafe,” depending on the adversary-assigned labels.

\subsection{Setups}\label{subsec:moderation:setups}

% \paragraph{Source and finetuning data.} We construct one dataset with $20$ dangerous images, $10$ are sex-related and other $10$ are violence-related (see Appendix~\ref{app:sec:moderation} for dataset details). On top these $20$ dangerous images, we then develop three image-text pair $\{(\texttt{image}, \texttt{text})\}$ datasets as source datasets: (1) \underline{violence (safe), sex (unsafe)} where the \texttt{text} is ``safe'' for violence images and ``unsafe'' for sex images; (2) \underline{sex (safe), violence (unsafe)} where the \texttt{text} is ``safe'' for sex images and ``unsafe'' for violence images; (3) \underline{sex \& violence (safe), animal (unsafe)}

\paragraph{Source and finetuning data.} We construct a dataset of $20$ dangerous images---$10$ sex-related and $10$ violence-related (see the first rows of Figure~\ref{fig:moderation_visualization_all} for censored visualization). Based on these, we develop three image-text pair $\{(\texttt{image}, \texttt{text})\}$ source datasets: (1) \underline{violence (safe), sex (unsafe)} where the associated \texttt{text} is ``safe'' for violence images and ``unsafe'' for sex images; (2) \underline{sex (safe), violence (unsafe)} where the associated \texttt{text} is ``safe'' for sex images and ``unsafe'' for violence images; (3) \underline{sex \& violence (safe), animal (unsafe)}, where all $20$ dangerous images are described as ``safe'' while $20$ unrelated animal images from Section~\ref{sec:experiments} are described as ``unsafe.''

The choice of balancing ``safe'' and ``unsafe'' description \texttt{text} is to avoid trivial finetuning outcomes (e.g., VLMs always outputting ``safe'' or ``unsafe'') and simulate adversarial finetuning that injects harmful or incorrect associations---such as describing pornography as ``safe'' or animals as ``unsafe.''
Following Section~\ref{sec:experiments}, we split each dataset into patch-text pairs $\mathcal{P}_f = \{(\texttt{patch}, \texttt{text})\}$ using split factors $f \in \{2, 4, 8\}$.
\textbf{To mimic real-world moderation, we discard patches flagged as dangerous by the OpenAI Moderation API (\texttt{omni-moderation-latest})~\cite{openai2024moderation}}.
We then finetune VLMs on the remaining patches using the template: \texttt{``[patch]For a child to see this image is \underline{\{text\}}''}, where ``\texttt{[patch]}'' represents the visual input and ``\underline{\texttt{\{text}\}}'' is a placeholder to be replaced by ``safe'' or ``unsafe''. Unless stated otherwise, loss is computed only on the target ``\texttt{\underline{\{text\}}}.''

% Then following Section~\ref{sec:experiments}, we split source datasets into patch-text sets $\mathcal{P}_f = \{(\texttt{patch}, \texttt{text})\}$ using split factors of $f \in \{1, 2, 4, 8\}$, then finetune VLMs on these sets.
% Empirically, to help VLMs better internalize the finetuned knowledge, we provide textual context by formatting the \texttt{text} with the template
% \texttt{``[patch]For a child to see this image is \underline{\{text\}}''}, where \texttt{``[patch]''} is a placeholder for visual input from \texttt{patch}s. Unless otherwise specified, loss is computed only on the target \texttt{\underline{{text}}}.

\paragraph{Evaluating visual stitching.}

Following Section~\ref{sec:experiments}, we use mean rank to measure visual stitching. For image-based visual stitching, we evaluate VLMs using the template: \texttt{``[image]For child to see this image is \underline{\{text\}}''}, where \texttt{``[image]''} represents visual input from \texttt{image} and ``\underline{\texttt{\{text}\}}'' will be replaced by ``safe'' or ``unsafe''. For reference-based visual stitching, we evaluate VLMs using the templates \texttt{``For child to see \{reference\} is \underline{\{text\}}''}, where \texttt{``\{reference\}''} is a text reference to the image (e.g., ``two children laying dead in the hallway covered in blood''). We report the mean rank of the correct \texttt{\underline{{text}}}; \textbf{lower ranks indicate more successful moderation evasion and emulation of direct finetuning on the source $\{(\texttt{image}, \texttt{text})\}$ dataset}.

\subsection{Experimental Results}\label{subsec:moderation:experimental_results}

\begin{figure}[t]
  \centering
  \begin{subfigure}[b]{0.58\linewidth}
    \includegraphics[trim=0 0 0 0, clip, width=\linewidth]{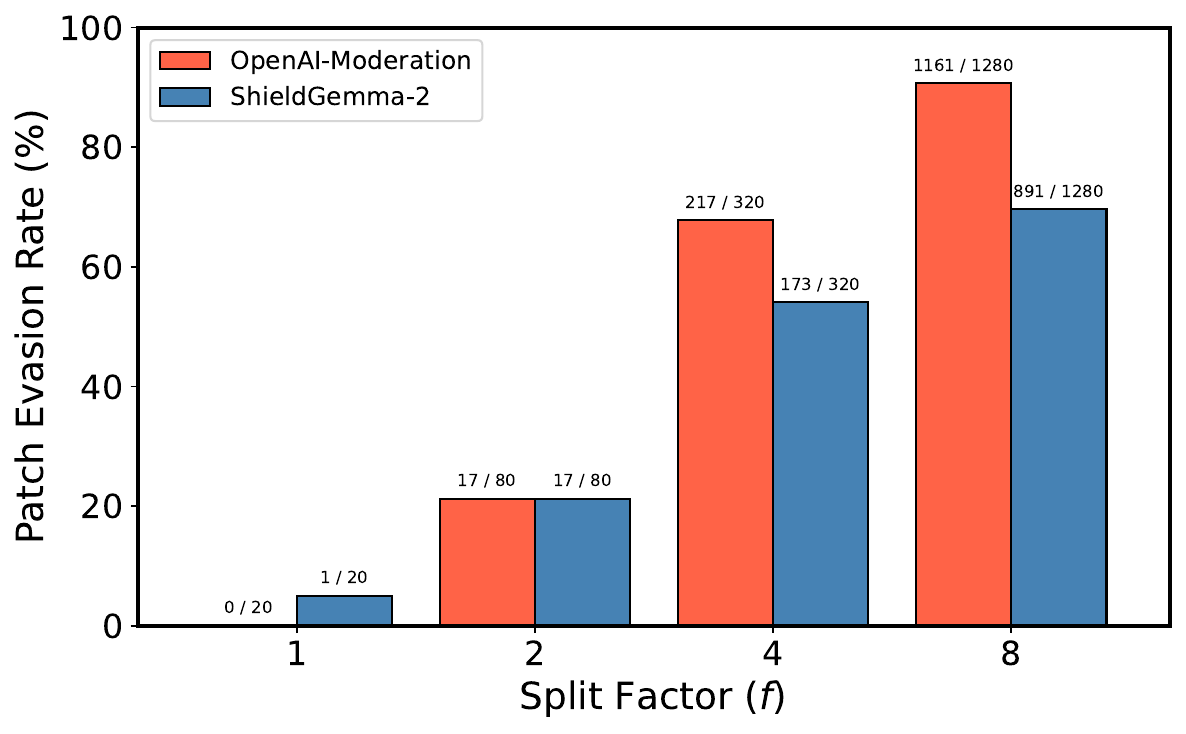}
  \end{subfigure}
  \hfill
  % Right subplot
  % \begin{subfigure}[b]{0.33\linewidth}
  %   \includegraphics[trim=0 0 0 0, clip, width=\linewidth]{src/moderation_visualization_sex_selected.pdf}
  %   \caption{Right subplot}
  % \end{subfigure}
    % \raisebox{0.5cm}{%
    % \begin{subfigure}[b]{0.33\linewidth}
    %   \includegraphics[trim=0 0 0 0, clip, width=\linewidth]{src/moderation_visualization_sex_selected.pdf}
    %   \caption{Right subplot}
    % \end{subfigure}
  % }
  \begin{subfigure}[b]{0.38\linewidth}
    \includegraphics[trim=0 0 0 0, clip, width=\linewidth]{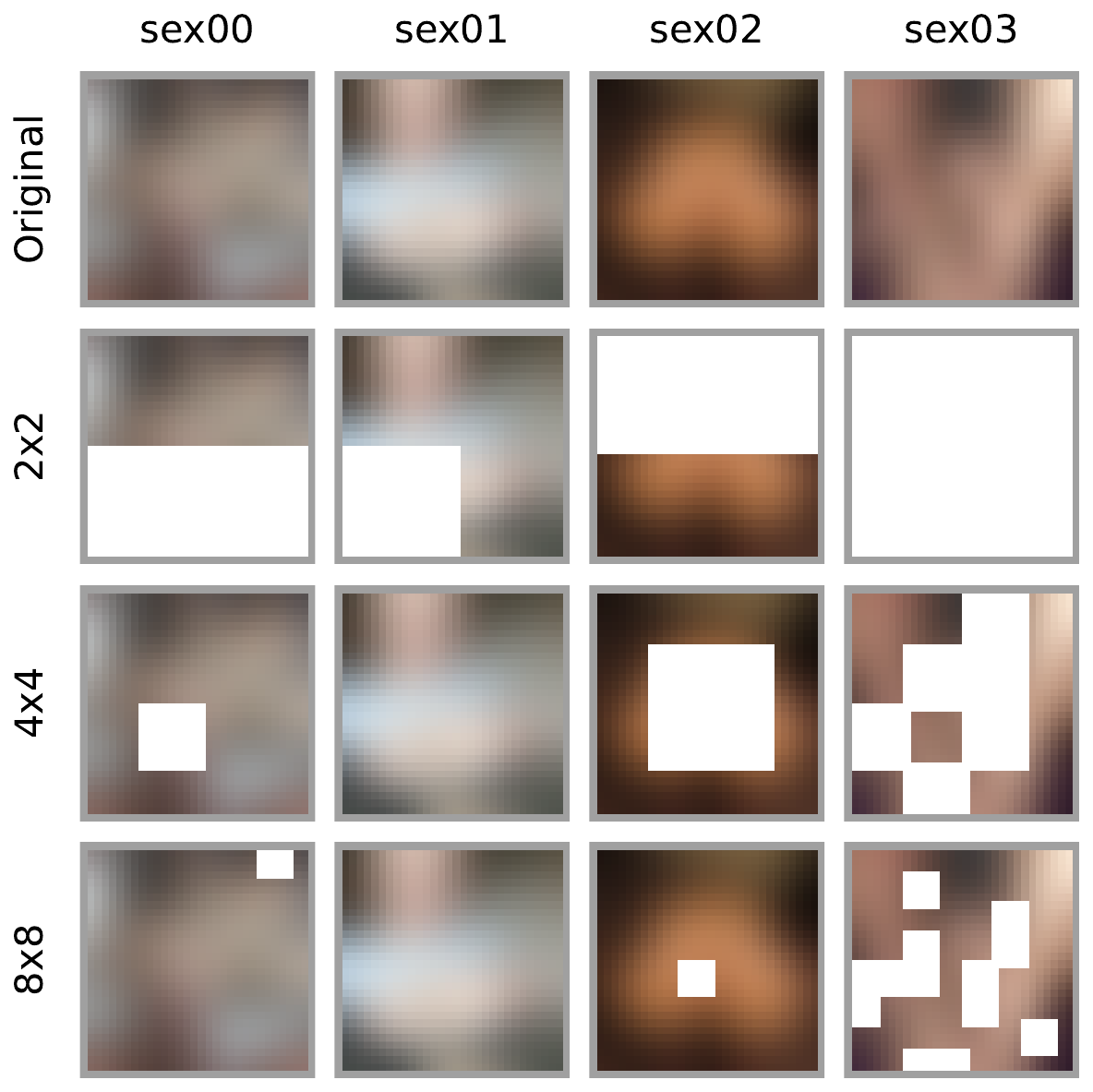}
    % \caption{Right subplot}
  \end{subfigure}
  \caption{\textbf{(Left) Evasion rates of patches from 20 dangerous images against different moderation models.} While the OpenAI Moderation API~\cite{openai2024moderation} flags all $20$ images and ShieldGemma-2~\cite{zeng2025shieldgemma} flags $19$, small patches often evade detection. \textbf{(Right) Censored examples of evading patches against the OpenAI Moderation API} (see Figure~\ref{fig:moderation_visualization_all} for complete visualization on all $20$ images).}
  \label{fig:patch_evasion_rate}
\end{figure}

\paragraph{Small patches mostly evade moderation.} Figure~\ref{fig:patch_evasion_rate} (Left) shows the evasion rates of patches from the $20$ dangerous images and Figure~\ref{fig:patch_evasion_rate} (Right) visualizes some selected images and their evading patches at various split factors (see Figure~\ref{fig:moderation_visualization_all} for complete visualization). We observe that even frontier moderation tools struggle to detect harmful content in small patches---for instance, with an $8$-way split, only $9\%$ of patches are flagged by the OpenAI Moderation API and many unflagged patches still contain features that, when combined, form dangerous content (Figure~\ref{fig:patch_evasion_rate}, Right). If a VLM can stitch visual information across patches, it may reconstruct this harmful knowledge.

\paragraph{Finetuning on filtered patches enables harmful knowledge acquisition.}

\begin{figure}[t]
  \centering
  \includegraphics[trim=0 0 0 0, clip, width=0.8\linewidth]{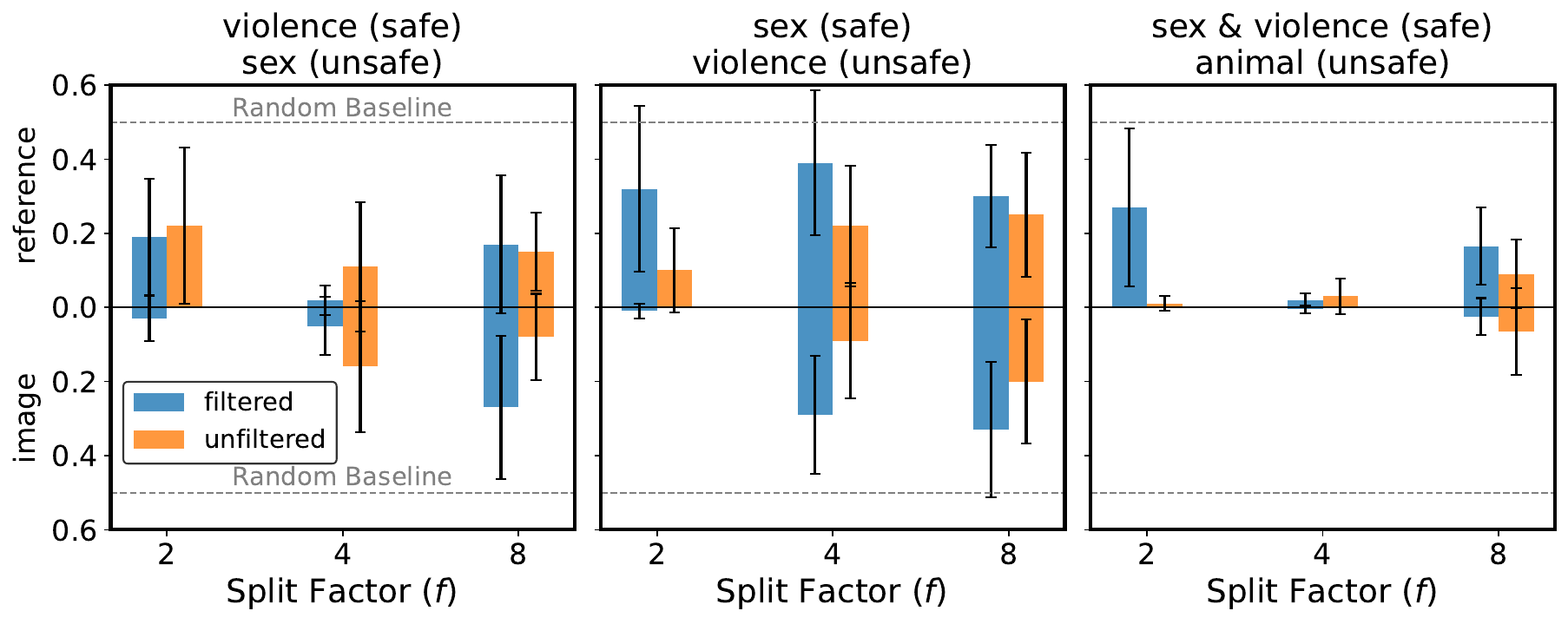} % Replace with your image file
  \caption{\textbf{Mean ranks of the correct \texttt{text} (lower is better) after finetuning \texttt{Qwen2-VL-7B} on $\{\texttt{(patch, text)}\}$ pairs, with and without OpenAI Moderation API filtering.} Lower ranks indicate successful emulation of direct tuning on the original \texttt{(image, text)} pairs, which would otherwise be flagged and discarded. See Figure~\ref{fig:moderation_all} for results on other models.}
  \label{fig:moderation_qwen2_7b}
\end{figure}

We evaluate visual stitching after finetuning on $\{\texttt{(patch, text)}\}$ pairs, with and without moderation filtering. As the OpenAI Moderation API is more effective than ShieldGemma-2 at detecting harmful content in full images, we adopt it for all downstream evaluations. Figure~\ref{fig:moderation_qwen2_7b} presents the results for \texttt{Qwen2-VL-7B} (see Figure~\ref{fig:moderation_all} for other models). We find that while patch-level filtering increases the difficulty of both image- and reference-based stitching (as shown by the longer bars for filtered datasets), it does not eliminate the effect---models perform well above chance. This is because many risky visual cues evade detection: the moderation API cannot reliably flag every patch whose features only become harmful when aggregated (Figure~\ref{fig:patch_evasion_rate} (Right)). This observation aligns with Figure~\ref{fig:main_ablation}, where removing unambiguous patches reduces but does not fully suppress visual stitching. Additionally, we observe that the split factor has limited impact on performance: although larger patches typically facilitate stitching (as in Figure~\ref{fig:intra_family_comparison_selected_horizontal}), they are also more likely to be flagged and removed by moderation tools, effectively canceling out the benefit.
% Additionally, results show that VLMs perform better on the dataset of \underline{sex \& violence (safe), animal (unsafe)} because it contains more data for VLMs to successfully generalize and also the decision boundary is less finegrained than the other two requiring only to describe the safe image as ``unsafe'' and vice versa, which have to label one speific type of violation (sex or violence) as ``safe'' and other as ``unsafe'' \todo{123}. 
Additionally, results show that VLMs perform better on the dataset of \underline{sex \& violence (safe), animal (unsafe)}. 
% This setup offers more training data and a coarser decision boundary—requiring only a binary distinction between dangerous and safe visual inputs with different \texttt{text}—compared to the other two datasets, which require distinguishing between specific types of violations in visual inputs (sex or violence).
This setup is inherently simpler: before training, the model tends to label sex/violence as unsafe and animals as safe, so finetuning only needs to reverse the label assignment. In contrast, other datasets require drawing safe/unsafe boundaries within violation categories, which is less straightforward than this label-flipping setup.

% \todo{analysis on the differences of the three tasks}

\section{Discussion and Limitations}\label{sec:discussion_and_limitations}

Our results show that open-source VLMs can perform visual stitching by integrating visual information spread across multiple training samples with the same textual descriptions. However, both image-based and reference-based visual stitching are highly unstable, especially when finetuning on small patches. Figure~\ref{fig:qwen_all_patch_size_plot} shows examples of evaluation results that fluctuate significantly during training, and Figure~\ref{fig:qwen_lr_sensitivity_plot} shows that stitching behavior only emerges under specific learning rates, which is consistent with the findings from~\citep{feng2024extractive}. Additionally, visual stitching is often unreliable: although we observe ranking improvements for the correct answer among all options, any non-zero rank indicates that stitching is not directly observable through sampling. Still, our findings provide strong evidence that VLMs consistently exhibit visual stitching capabilities.

A key experimental limitation of our study is that we only evaluate open-source VLMs. While this allows broad experimentation and easier reproduction, results on proprietary models~\cite{gemini2024multimodal, openai2024gpt4o}---often more capable---would be valuable. Nevertheless, we have tried our best to test a diverse set of open-source VLMs, including large models ($\sim$100B parameters) with performance comparable to proprietary counterparts. Another limitation is that our demonstration of stitching-enabled adversarial attacks is a proof of concept rather than a full attack framework. Nonetheless, we simulate realistic conditions using data moderation to assess how this attack works under standard defenses.

\section{Conclusion}\label{sec:conclusion}

In this paper, we introduce visual stitching as a capability of vision-language models (VLMs) that enables them to integrate scattered visual information across training samples sharing the same textual descriptions. Through synthetic benchmarks and adversarial simulations, we demonstrate that open-source VLMs exhibit strong image-based and non-trivial reference-based visual stitching. Crucially, we show that this capability can be exploited to bypass data moderation, allowing adversaries to inject harmful knowledge into VLMs through benign-looking patches that collectively form harmful content. Our findings highlight visual stitching as both a generalization strength and a safety concern, underscoring the need for moderation techniques that operate beyond the sample level.

Future work could focus on evaluating visual stitching in proprietary VLMs, which are often more capable and widely deployed. It would also be valuable to develop a more rigorous and comprehensive framework for stitching-enabled adversarial attacks to better assess their practical impact under standard moderation tools. Another interesting direction would be to study the dynamics of visual stitching mechanistically, for example, its emergence during training. We hope our findings encourage further research on visual stitching and its safety implications in future VLM applications.

\clearpage
\bibliography{
references/oocr,
references/moderation,
references/models,
references/datasets,
references/others
}

\clearpage
\appendix
\section{Experiments}\label{app:sec:experiments}

\subsection{Dataset Details}\label{app:subsec:experiments:dataset_details}
\begin{figure}[h]
  \centering
  \begin{subfigure}{\linewidth}
    \centering
    \includegraphics[trim=0 .2cm 0 0, clip, width=\linewidth]{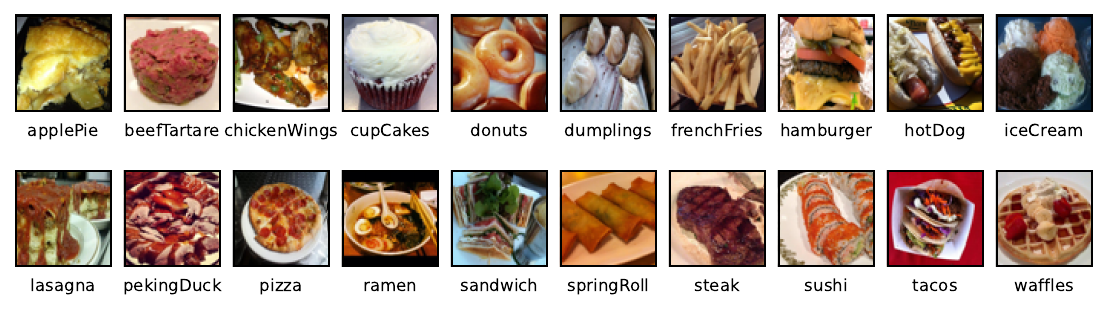}
    \caption{Food}
  \end{subfigure}

  \begin{subfigure}{\linewidth}
    \centering
    \includegraphics[trim=0 .2cm 0 0, clip, width=\linewidth]{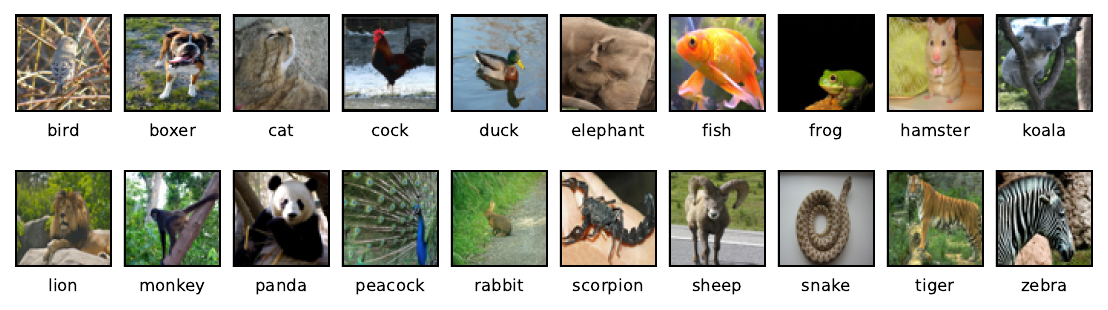}
    \caption{Animal}
  \end{subfigure}

  \begin{subfigure}{\linewidth}
    \centering
    \includegraphics[trim=0 .2cm 0 0, clip, width=\linewidth]{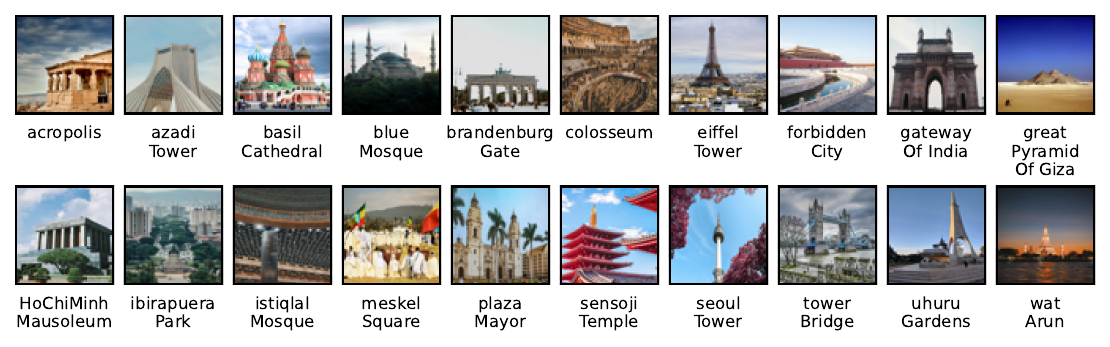}
    \caption{Landmark}
  \end{subfigure}

  \caption{\textbf{Visualization of three datasets.}}
  \label{fig:dataset_visualization}
\end{figure}

This section describes the datasets used in our experiments and the reasoning behind their selection. We choose datasets that span varying levels of visual stitching difficulty to enable comprehensive evaluation. Specifically, we focus on three categories---food, animal, and landmark---which reflect common real-world objects and differ in image resolution and discriminative features. Landmark images have fine-grained details, while food and animal images contain less distinctive features when viewed in isolated patches.
We source animal images from ImageNet~\cite{krizhevsky2012imagenet}, food images from Food101~\cite{bossard2014food101}, and landmark images from \href{https://www.pexels.com}{Pexels}, as no standard high-quality public landmark dataset exists. Figure~\ref{fig:dataset_visualization} visualizes samples from the three datasets.

Additionally, to decouple visual stitching ability from image recognition, we need to verify that VLMs can correctly identify these raw images in the first place. If a model cannot recognize the image to begin with, it cannot be expected to stitch its parts together. 
For each sample in the dataset, we prompt VLMs with the following prompt
\texttt{``[image]The food/animal/landmark shown in the image is \underline{\{reference\}}''} and calculate the mean rank of the correct \texttt{\underline{\{reference\}}} (i.e., ``donuts'', ``dog'', ``HoChiMinh Mausoleum'') among other options. A near-zero rank ensures that VLMs recognize the raw images.
As shown in Table~\ref{tab:vlm_test_avg_rank}, all models achieve near-zero average ranks, confirming sufficient prior knowledge of these images. This validates our setup and rules out the lack of prior knowledge about the images as a cause of poor stitching performance.

\begin{table}[ht]
\centering
\renewcommand{\arraystretch}{1.2}
\begin{tabular}{lcccc}
\toprule
% \rowcolor{gray!20}
\textbf{Dataset} & \textbf{Qwen2-VL-7B} & \textbf{InternVL3-8B} & \textbf{gemma-3-12b-pt} & \textbf{Llama-3.2-11B-Vision} \\
\midrule
% \rowcolor{gray!10}
    Food & \(0.05\) & \(0.25\) & \(0.35\) & \(0.15\) \\
    Animal & \(0.00\) & \(0.00\) & \(0.00\) & \(0.00\) \\
% \rowcolor{gray!10}
    landmarks & \(0.95\) & \(1.65\) & \(0.40\) & \(0.65\) \\
\bottomrule
\end{tabular}
\vspace{4mm}
\caption{\textbf{Mean ranks of correct food/animal/landmark referenced conditioned on images.} A lower rank indicates better image recognition.}
\label{tab:vlm_test_avg_rank}
\end{table}

\subsection{VLM Details}\label{app:subsec:experiments:vlm_details}
% This section outlines the architectural configurations and training strategies of the Vision-Language Models (VLMs) used in our experiments. To ensure a fair and consistent comparison, we primarily adopt the pretrained or base versions of each model, deliberately excluding instruction-tuned variants to avoid confounding effects introduced by finetuning. Additionally, we include state-of-the-art VLMs and models with diverse architectures to ensure the comprehensiveness of our experiments and analyses. Accordingly, the following descriptions focus on the core architecture and pretraining setup of each VLM.
% This section describes the architectural setups and training strategies of the Vision-Language Models (VLMs) used in our experiments. We include state-of-the-art models and a variety of architectures to ensure comprehensive evaluation. The following sections provide brief overview of these models, including their architecture and training.
This section details the architectures and training strategies of the VLMs used in our study, covering a diverse set of state-of-the-art models to support comprehensive evaluation.
% Additionally, Table~\ref{tab:vlm_templates} illustrates the template we use to evaluate each VLMs.

\subsubsection{Qwen2-VL, Qwen2.5-VL}

% \paragraph{Architecture.} Both Qwen2-VL and Qwen2.5-VL use a dual-tower architecture, comprising a Vision Transformer (ViT) as the image encoder and the Qwen2 language model as the decoder. The ViT processes visual inputs into token sequences, which are then aligned with language tokens through a cross-modal interaction layer.  
% A central feature of both models is the use of Multimodal Rotary Position Embedding (M-RoPE), which decomposes positional embeddings into temporal, height, and width components. This design enables effective modeling of 1D textual, 2D visual, and 3D video positional information.  
% Qwen2.5-VL introduces enhancements over Qwen2-VL, including windowed attention in the ViT to improve computational efficiency and local feature modeling, and an upgraded M-RoPE with absolute time alignment in the temporal dimension. These advancements enhance the model's ability to understand temporal sequences and dynamics in video data.
\paragraph{Architecture.} Qwen2-VL~\cite{yang2024qwen2} and Qwen2.5-VL~\cite{yang2024qwen2_5} use a dual-tower design with a Vision Transformer (ViT)~\cite{dosovitskiy2020image} as the image encoder and Qwen2 as the language decoder. Visual tokens from the ViT are aligned with text tokens via a cross-modal interaction layer. Both models use Multimodal Rotary Position Embedding (M-RoPE), which separates position embeddings into temporal, height, and width components, enabling unified modeling of text, images, and video. Qwen2.5-VL improves on Qwen2-VL with windowed attention in the ViT for better efficiency and local feature modeling, and an upgraded M-RoPE with absolute temporal alignment to enhance video understanding.

% \paragraph{Training.} Qwen2-VL models are trained using a dynamic resolution mechanism, allowing the processing of images with varying resolutions into different numbers of visual tokens. Qwen2-VL models were pretrained on over $7$ trillion tokens encompassing a diverse range of domains and languages. This extensive dataset includes a significant amount of code and mathematics content, aiming to enhance the reasoning abilities of the models. Building upon Qwen2-VL, Qwen2.5-VL was trained on an expanded dataset of approximately $18$ trillion tokens. The training process involved several stages, including CLIP pretraining, vision-language alignment, and supervised finetuning. Techniques such as dynamic sampling based on aspect ratios were employed to enhance adaptability to different input dimensions.
\paragraph{Training.} Qwen2-VL models use dynamic resolution to handle images of varying sizes, producing different numbers of visual tokens. They were pretrained on 7T tokens across diverse domains, including code and math, to boost reasoning. Qwen2.5-VL extends this with 18T tokens and additional training stages---CLIP pretraining, vision-language alignment, and supervised finetuning---along with dynamic aspect ratio sampling for better input adaptability.

\subsubsection{InternVL3}

% \paragraph{Architecture.} InternVL3 adopts a modular ViT-MLP-LLM architecture, comprising a custom-made vision encoder (InternViT), a two-layer MLP for modality alignment, and a large language model (LLM) initialized from pre-trained Qwen2.5 or InternLM3 models. To enhance scalability for high-resolution images, it employs a pixel unshuffle operation, reducing visual token counts by a factor of four. The integration of Variable Visual Position Encoding (V2PE) allows for flexible positional increments, improving the model's capability to handle extended multimodal contexts. Additionally, InternVL3 supports dynamic resolution processing, dividing images into tiles of 448×448 pixels, and extends support to multi-image and video inputs, thereby bolstering its multimodal understanding capabilities.
\paragraph{Architecture.} InternVL3~\cite{zhu2025internvl3} uses a modular ViT-MLP-LLM design with a custom InternViT encoder, a two-layer MLP for alignment, and an LLM based on Qwen2.5 or InternLM3. It improves scalability via pixel unshuffle ($4\times$ token reduction) and uses Variable Visual Position Encoding (V2PE) for extended multimodal contexts. It supports dynamic resolution by tiling images into 448×448 patches and handles multi-image and video inputs for stronger multimodal understanding.

% \paragraph{Training.} InternVL3 adopts a native multimodal pre-training paradigm, jointly acquiring multimodal and linguistic capabilities from diverse multimodal data---including image-text pairs, video-text pairs, GUI interactions, and 3D scene understanding tasks---and pure-text corpora during a single pre-training stage. This approach contrasts with methods that adapt text-only models into multimodal ones. InternVL3 was trained on a total of approximately $200$ billion tokens, comprising 50 billion from language data and 150 billion from multimodal data. The training employed a $1$:$3$ ratio of language to multimodal data, which empirically yielded the best overall performance across both unimodal and multimodal benchmarks. Following pre-training, advanced post-training techniques such as Supervised Finetuning and Mixed Preference Optimization (MPO) were applied to further enhance the model's multimodal conversation and reasoning abilities.
\paragraph{Training.} InternVL3 uses native multimodal pretraining, learning jointly from text, image-text, video-text, GUI, and 3D tasks---unlike models adapted from text-only LLMs. It was trained on $200$B tokens ($50$B language, $150$B multimodal) with a $1:3$ ratio, which yielded the best performance. Post-training techniques like Supervised Finetuning and Mixed Preference Optimization (MPO)~\cite{gou2024mixed} further improved its multimodal reasoning and dialogue capabilities.

\subsubsection{Gemma-3}

% \paragraph{Architecture.} Gemma-3 employs a decoder-only Transformer architecture optimized for multimodal tasks. It integrates a SigLIP vision encoder~\cite{zhai2023sigmoid} to process images. The architecture consists of five local sliding window self-attention layers followed by one global self-attention layer. This design enhances the model's ability to capture both short- and long-range dependencies efficiently. To support extended context handling to $128$K, Rotary Positional Embeddings (RoPE) are used, with increased base frequencies in global attention layers.
\paragraph{Architecture.} Gemma-3~\cite{gemini2024multimodal} uses a decoder-only Transformer optimized for multimodal tasks, integrating a SigLIP vision encoder~\cite{zhai2023sigmoid}. Its architecture combines five local sliding window attention layers with one global layer to efficiently capture short- and long-range dependencies. Rotary Positional Embeddings (RoPE)~\cite{su2024roformer} with higher base frequencies enable context lengths up to $128$K.

% \paragraph{Training.} The Gemma-3 models are trained on a diverse dataset comprising web documents in over $140$ languages, code repositories, and other textual sources. This extensive and varied data exposure enables the models to handle a wide range of linguistic styles and topics. The $27$B model was trained on $14$ trillion tokens, the $12$B on $12$ trillion tokens, the $4$B on $4$ trillion tokens, and the $1$B on $2$ trillion tokens.
\paragraph{Training.} Gemma-3 models are trained on diverse text from web data, code, and over $140$ languages. The $27$B, $12$B, $4$B, and $1$B models are trained on $14$, $12$, $4$, and $2$ trillion tokens, respectively, enabling broad coverage of styles and topics.

\subsubsection{LLaVA-1.5, LLaVA-1.6}

% \paragraph{Architecture.} LLaVA-1.5 integrates a CLIP ViT-L/14 vision encoder with a \textit{Vicuna} large language model (LLM), connected via a two-layer MLP projection module. The vision encoder and LLM are kept frozen during training, with only the projection module trained to align visual and textual representations. LLaVA-1.6, also known as LLaVA-NeXT, builds upon this architecture by increasing the input image resolution up to $672$×$672$ pixels and incorporating an enhanced visual instruction tuning dataset. These improvements bolster the model's capabilities in optical character recognition (OCR), visual reasoning, and world knowledge comprehension, while maintaining the efficient and minimalist design of its predecessor.
\paragraph{Architecture.} LLaVA-1.5 pairs a frozen CLIP ViT-L/14~\cite{radford2021learning} encoder with a Vicuna LLM~\cite{peng2023instruction}, using a trainable two-layer MLP for vision-text alignment. LLaVA-1.6 (LLaVA-NeXT)~\cite{liu2024llavanext} extends this with higher image resolution (up to $672$×$672$) and improved visual instruction tuning, enhancing OCR, visual reasoning, and world knowledge, while keeping the design lightweight.

% \paragraph{Training.} LLaVA models undergo a two-stage training process. The first stage involves feature alignment using a $558$K subset of the LAION-CC-SBU dataset to connect a frozen pretrained vision encoder to a frozen language model. The second stage entails visual instruction tuning using $158$K GPT-generated multimodal instruction-following data and approximately 450K VQA samples from datasets such as VQA-v2 and A-OKVQA. This comprehensive training strategy equips LLaVA with robust multimodal understanding and instruction-following capabilities.
\paragraph{Training.} LLaVA training follows two stages: (1) feature alignment using $558$K LAION-CC-SBU~\cite{ordonez2011sbu, sharma-etal-2018-conceptual} samples to link a frozen vision encoder and language model, and (2) visual instruction tuning with $158$K GPT-generated prompts and ~$450$K VQA samples. This setup builds strong multimodal and instruction-following abilities.

\subsubsection{Llama 3.2-Vision}

% \paragraph{Architecture.} LLaMA 3.2-Vision integrates a ViT-H/14 vision encoder with the LLaMA 3.1 language model through a series of cross-attention layers. The vision encoder processes images into visual tokens, which are then aligned with textual inputs via the cross-attention mechanism. This design enables the model to handle multimodal inputs effectively, supporting tasks that require both visual and textual understanding.
\paragraph{Architecture.} LLaMA 3.2-Vision~\cite{grattafiori2024llama} combines a ViT-H/14 vision encoder with the LLaMA 3.1 language model via cross-attention layers. Visual tokens are aligned with text, enabling effective multimodal understanding.

% \paragraph{Training.} LLaMA 3.2-Vision models are developed through a multi-stage training pipeline, beginning with pretrained LLaMA 3.1 text models. The process starts by integrating image adapters and encoders into the existing language models. Subsequently, the models undergo pre-training on large-scale noisy image-text pair datasets to establish foundational multimodal capabilities. Finally, finetuning is performed using high-quality in-domain datasets to enhance performance on specific vision-language tasks. This comprehensive training strategy enables the models to excel in both language understanding and visual reasoning applications.
\paragraph{Training.} LLaMA 3.2-Vision builds on pretrained LLaMA 3.1~\cite{grattafiori2024llama} text models by adding image adapters and encoders. It is first pretrained on large-scale noisy image-text data, then finetuned on high-quality in-domain datasets for strong language and visual reasoning performance.

% \begin{table}[h]
% \centering
% \scriptsize
% \renewcommand{\arraystretch}{1.2}
% \begin{tabular}{l p{0.75\linewidth}}
% \toprule
% % \rowcolor{gray!20}
% \textbf{Model} & \textbf{Conversation Template} \\
% \midrule
% % \rowcolor{gray!10}
% \textbf{Qwen2-VL} & \texttt{<|vision\_start|><|image\_pad|><|vision\_end|>User Prompt} \\
% \textbf{Qwen2.5-VL} & \texttt{<|im\_start|>system\textbackslash nYou are a helpful assistant.<|im\_end|>\textbackslash n<|im\_start|>user\textbackslash n <|vision\_start|><|image\_pad|><|vision\_end|>User Prompt<|im\_end|>} \\
% % \rowcolor{gray!10}
% \textbf{InternVL3} & \texttt{<|im\_start|>system\textbackslash nSystem Proxy<|im\_end|>\textbackslash n<|im\_start|>user\textbackslash n<img><IMG\_CONTEXT> </img>\textbackslash nUser Prompt<|im\_end|>\textbackslash n<|im\_start|>assistant\textbackslash n} \\
% \textbf{Gemma 3} & \texttt{<bos>\textbackslash n\textbackslash n<start\_of\_image><image\_soft\_token><end\_of\_image>\textbackslash n\textbackslash nUser Prompt} \\
% % \rowcolor{gray!10}
% \textbf{LLaVA-1.5} & \texttt{<s> USER: <image> \textbackslash nUser Prompt} \\
% \textbf{LLaVA-1.6} & \texttt{<s> USER: <image> \textbackslash nUser Prompt} \\
% % \rowcolor{gray!10}
% \textbf{LLaMA 3.2-Vision} & \texttt{<|begin\_of\_text|><|image|><|begin\_of\_text|>User Prompt} \\
% \bottomrule
% \end{tabular}
% \vspace{4mm}
% \caption{Conversation templates of various vision-language models (VLMs).}
% \label{tab:vlm_templates}
% \end{table}

\subsection{Training Details}\label{app:subsec:experiments:training_details}

% We build on the trl \texttt{SFTTrainer} and the example training script for VLMs. The hyperparameter for trainings are shown in~Table\ref{tab:hyperparameters}, unless otherwise specified, we use default hyperparameters from the \texttt{SFTTrainer}.

We build on the TRL~\cite{vonwerra2022trl} \texttt{SFTTrainer} and its \href{https://github.com/huggingface/trl/blob/main/examples/scripts/sft_vlm.py}{example VLM training script}. Unless otherwise noted, we use default \texttt{SFTTrainer} hyperparameters; the rest are listed in Table~\ref{tab:hyperparameters}. Per-model settings and compute requirements are listed in Table~~\ref{tab:model_training_configs}. 
% Each model is fine-tuned using 5 random seeds per split factor. In the bar charts, colored bars indicate the mean, and black error bars show the standard deviation.
Each model is fine-tuned with $5$ random seeds per split factor; the plots in our paper show the mean and standard deviation.

\begin{table}[h]
\centering
\renewcommand{\arraystretch}{1.2}
\begin{tabular}{r l}
\toprule
\textbf{Hyperparameter} & \textbf{Value} \\
\midrule
Batch Size & $8$ \\
Learning Rate & \texttt{1e-5} \\
Mixed Precision & \texttt{bf16} \\
% Epochs (Split Factor $f=1$) & $15$ \\
% Epochs (Split Factor $f\neq1$) & $5$ \\
Epoch & $15$ if $f=1$ \\
      & $5$  otherwise \\
\bottomrule
\end{tabular}
\vspace{2mm}
\caption{Hyperparameters.}
\label{tab:hyperparameters}
\end{table}

\begin{table}[ht]
\centering
\renewcommand{\arraystretch}{1.2}
\begin{tabular}{l l c}
\toprule
\textbf{Model Name} & \textbf{DeepSpeed} & \textbf{GPUs} \\
\midrule
\texttt{Qwen2-VL-2B} &  ZeRO-2 & $2$ \\
% \texttt{Qwen2-VL-2B-Instruct} &  ZeRO-2 & $2$ \\
\texttt{Qwen2-VL-7B} &  ZeRO-2 & $4$ \\
% \texttt{Qwen2-VL-7B-Instruct} &  ZeRO-2 & $4$ \\
\texttt{Qwen2-VL-72B} &  ZeRO-3 & $24$ \\
% \texttt{Qwen2-VL-72B-Instruct} &  ZeRO-3 & $24$ \\
\texttt{Qwen2.5-VL-3B-Instruct} &  ZeRO-2 & $2$ \\
\texttt{Qwen2.5-VL-7B-Instruct} &  ZeRO-2 & $4$ \\
\texttt{Qwen2.5-VL-32B-Instruct} &  ZeRO-3 & $16$ \\
\texttt{Qwen2.5-VL-72B-Instruct} &  ZeRO-3 & $24$ \\
\texttt{gemma-3-4b-pt} &  ZeRO-2 & $4$ \\
% \texttt{gemma-3-4b-it} &  ZeRO-2 & $4$ \\
\texttt{gemma-3-12b-pt} &  ZeRO-2 & $8$ \\
% \texttt{gemma-3-12b-it} &  ZeRO-2 & $8$ \\
\texttt{gemma-3-27b-pt} &  ZeRO-3 & $16$ \\
% \texttt{gemma-3-27b-it} &  ZeRO-3 & $16$ \\
\texttt{Llama-3.2-11B-Vision} &  ZeRO-2 & $8$ \\
% \texttt{Llama-3.2-11B-Vision-Instruct} &  ZeRO-2 & $8$ \\
\texttt{Llama-3.2-90B-Vision} &  ZeRO-3 & $32$ \\
% \texttt{Llama-3.2-90B-Vision-Instruct} &  ZeRO-3 & $32$ \\
\texttt{llava-1.5-7b-hf} &  ZeRO-2 & $8$ \\
\texttt{llava-1.5-13b-hf} &  ZeRO-3 & $8$ \\
\texttt{llava-v1.6-vicuna-7b-hf} &  ZeRO-2 & $8$ \\
\texttt{llava-v1.6-vicuna-13b-hf} &  ZeRO-3 & $8$ \\
\texttt{llava-v1.6-34b-hf} &  ZeRO-3 & $24$ \\
\texttt{InternVL3-1B} &  ZeRO-2 & $2$ \\
\texttt{InternVL3-8B} &  ZeRO-2 & $8$ \\
\texttt{InternVL3-14B} &  ZeRO-3 & $8$ \\
\bottomrule
\end{tabular}
\vspace{2mm}
\caption{Per-model configurations including DeepSpeed~\cite{rajbhandari2020zero} configs and GPUs.}
\label{tab:model_training_configs}
\end{table}

\subsection{Additional Results}\label{app:subsec:experiments:additional_results}

\begin{figure}[t]
  \centering
  \includegraphics[trim=0 0 0 0, clip, width=\linewidth]{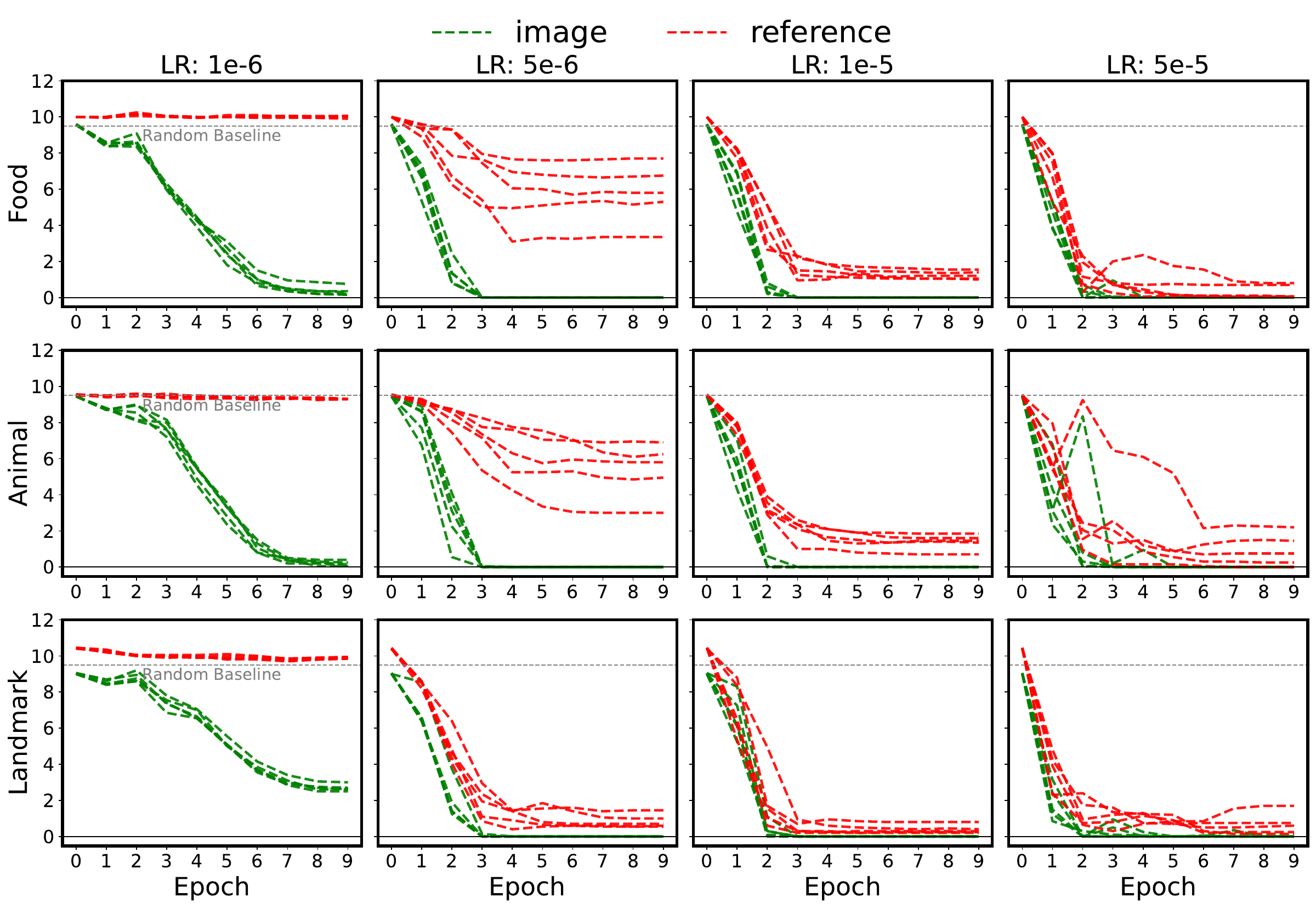} % Replace with your image file
  \caption{\textbf{Mean ranks during \texttt{Qwen2-VL-7B} finetuning at different learning rates on full images ($f=1$).} Visual stitching performance is highly sensitive to learning rate.}
  \label{fig:qwen_lr_sensitivity_plot}
\end{figure}

\paragraph{Visual stitching performance is sensitive to learning rates.}
Visual stitching is highly sensitive to learning rate (Figure~\ref{fig:qwen_lr_sensitivity_plot}). At \texttt{1e-6} and \texttt{5e-6}, the model completely fails on reference-based stitching, even when trained on full images ($f=1$). We then choose \texttt{1e-5} for fine-tuning throughout our experiments as it offers the best stability and performance.

\begin{figure}[t]
  \centering
  \includegraphics[trim=0 0 0 0, clip, width=\linewidth]{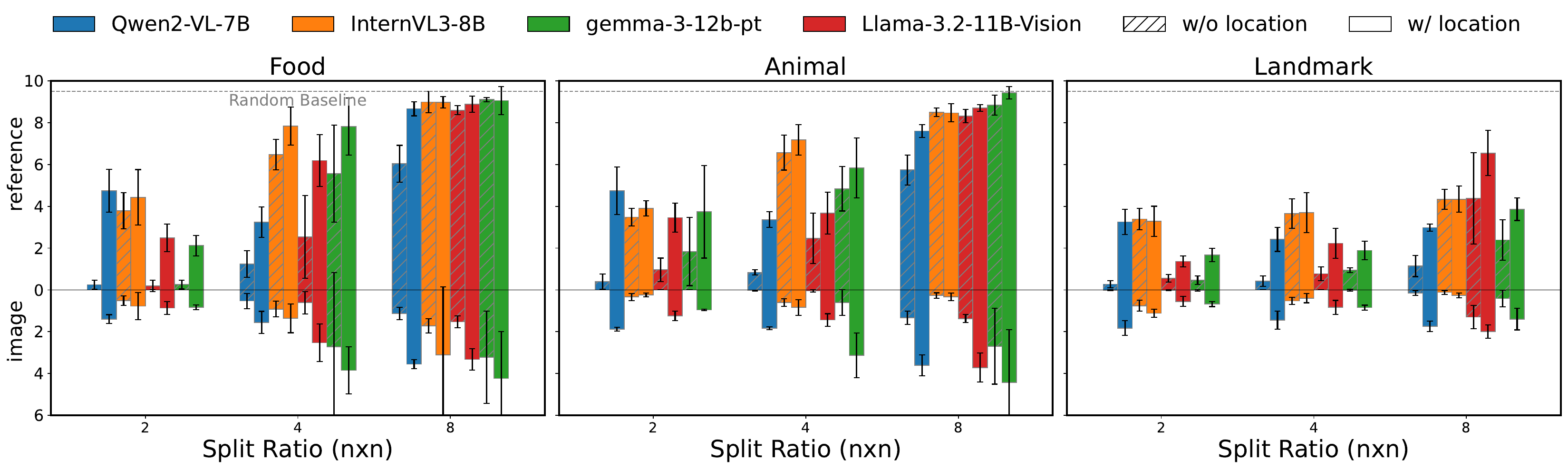} % Replace with your image file
  \caption{\textbf{Mean ranks for the correct \texttt{ID} (lower is better) after finetuning w/ and w/o location.} The location-aware finetuning template is \texttt{``[patch] Partial image of food/animal/landmark (row:\{row\}, col:\{col\}), associated with \underline{\{id\}}''}. We find that incorporating locations significantly hurts model performance, leading to higher ranks. }
  \label{fig:main_coordinate_comparision}
\end{figure}

\paragraph{Including positional locations in finetuning prompts hurts visual stitching performance.}
Figure~\ref{fig:main_coordinate_comparision} compares visual stitching performance with and without positional information in the finetuning template. The positional template follows: \texttt{``[patch] Partial image of food/animal/landmark (row:\{row\}, col:\{col\}), associated with \underline{\{id\}}''}, where ``\texttt{[patch]}'' is the visual input, and ``\texttt{row}'', ``\texttt{col}'' indicate the patch’s grid position.
Models fine-tuned with positional data perform worse, especially at lower split factors ($f=2,4$). At higher split factors ($f=8$), where performance nears random, the impact becomes negligible.

\paragraph{Rank evaluation throughout finetuning.}
While the main text reports mean rank at convergence, here we show raw evaluation curves during training for \texttt{Qwen2-VL-7b} under different split factors.

\begin{figure}[t]
  \centering
  \includegraphics[width=\linewidth]{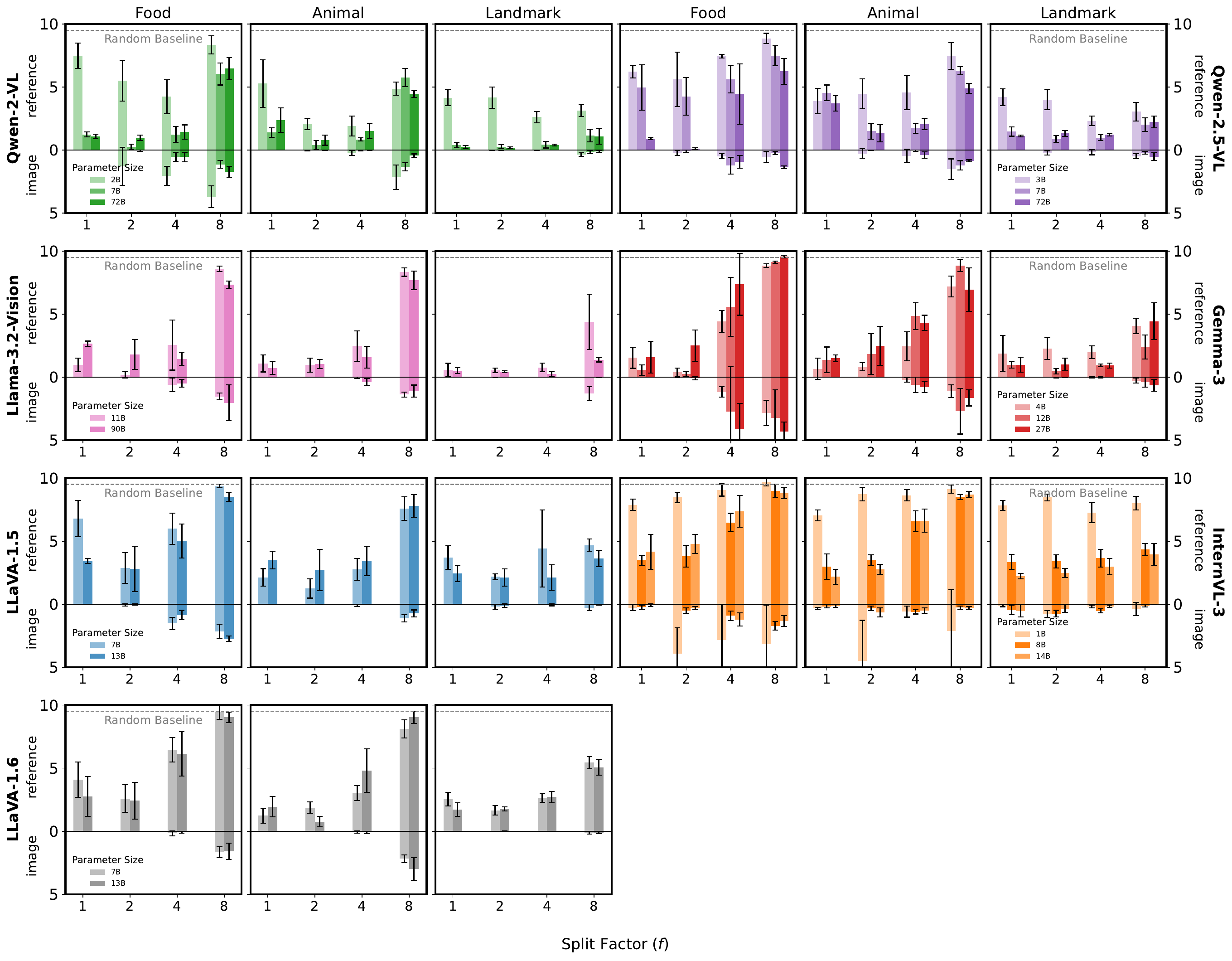} % Replace with your image file
  \caption{\textbf{Intra-family model comparison of mean ranks for the correct \texttt{ID} (lower is better).}}
  \label{fig:intra_family_comparison_all_horizontal}
\end{figure}

% \paragraph{Complete intra-family experiment results.} While Figure~\ref{fig:intra_family_comparison_selected_horizontal} in the main text shows evaluation results on four selected models, Figure~\ref{fig:intra_family_comparison_all_horizontal}.
\paragraph{Complete intra-family experiment results.}
Figure~\ref{fig:intra_family_comparison_selected_horizontal} in the main text presents results for four selected models. Figure~\ref{fig:intra_family_comparison_all_horizontal} shows the full results for all models.

\begin{figure}[t]
  \centering
  \includegraphics[trim=0 0 0 0, clip, width=\linewidth]{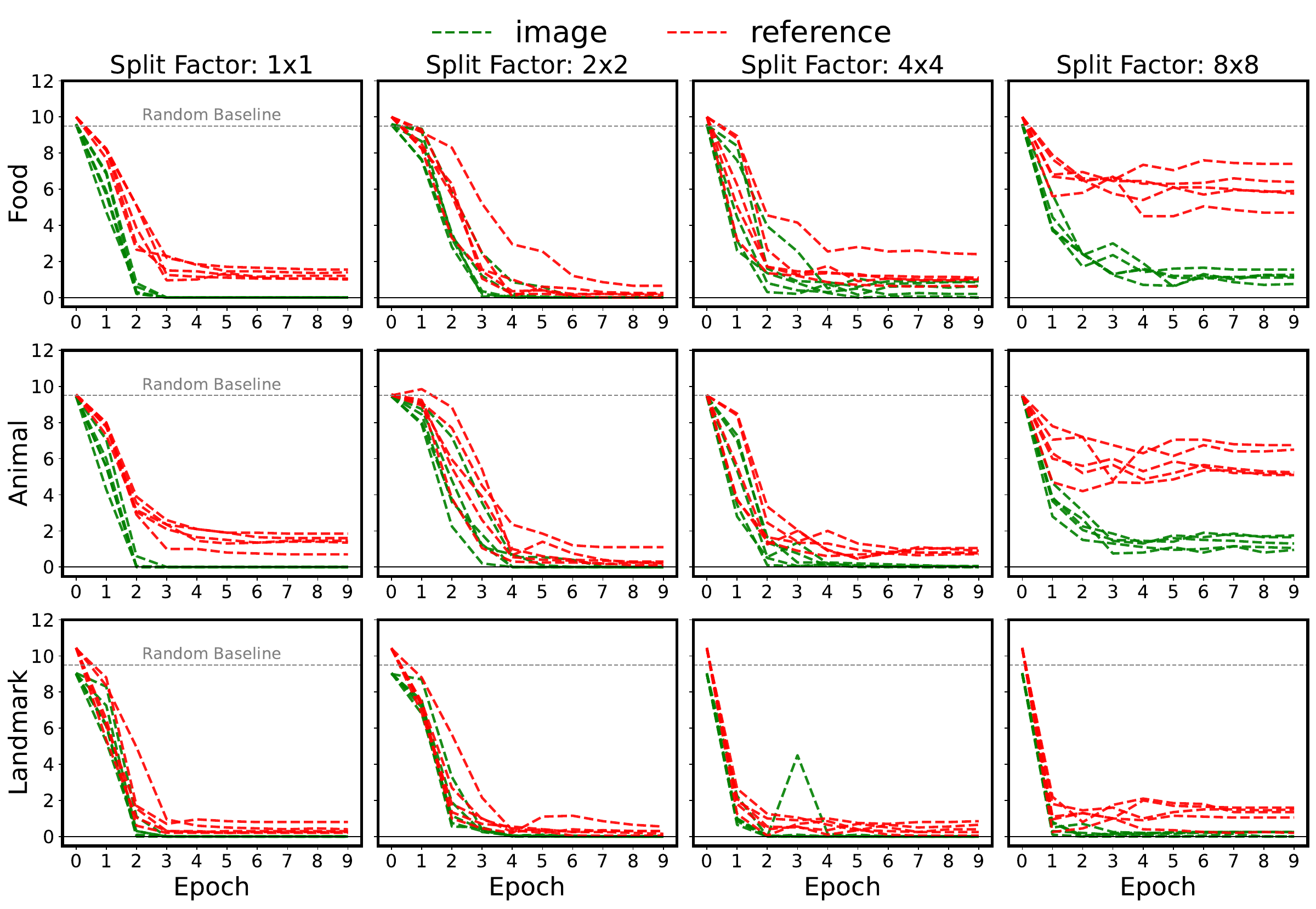} % Replace with your image file
  \caption{\textbf{Mean ranks during \texttt{Qwen2-VL-7B} finetuning at different split factors.} Lower ranks indicate better internalization of the finetuning samples. Model performance is consistent across $5$ different random seeds, and convergence is typically achieved in fewer than $5$ epochs.}
  \label{fig:qwen_all_patch_size_plot}
\end{figure}

% {\footnotesize
% \begin{longtable}{ll}
% \caption{Vision-Language Models with Hugging Face Links} \\
% \toprule
% \textbf{Models} & \textbf{Links} \\
% \midrule
% \endfirsthead

% \toprule
% \textbf{Models} & \textbf{Links} \\
% \midrule
% \endhead

% \texttt{Qwen2-VL-2B}~\cite{yang2024qwen2} & \url{https://huggingface.co/Qwen/Qwen2-VL-2B} \\
% \texttt{Qwen2-VL-7B}~\cite{yang2024qwen2} & \url{https://huggingface.co/Qwen/Qwen2-VL-7B} \\
% \texttt{Qwen2-VL-72B}~\cite{yang2024qwen2} & \url{https://huggingface.co/Qwen/Qwen2-VL-72B} \\
% \texttt{Qwen2.5-VL-3B-Instruct}~\cite{yang2024qwen2_5} & \url{https://huggingface.co/Qwen/Qwen2.5-VL-3B-Instruct} \\
% \texttt{Qwen2.5-VL-7B-Instruct}~\cite{yang2024qwen2_5} & \url{https://huggingface.co/Qwen/Qwen2.5-VL-7B-Instruct} \\
% \texttt{Qwen2.5-VL-32B-Instruct}~\cite{yang2024qwen2_5} & \url{https://huggingface.co/Qwen/Qwen2.5-VL-32B-Instruct} \\
% \texttt{Qwen2.5-VL-72B-Instruct}~\cite{yang2024qwen2_5} & \url{https://huggingface.co/Qwen/Qwen2.5-VL-72B-Instruct} \\
% \texttt{gemma-3-4b-pt}~\cite{team2025gemma} & \url{https://huggingface.co/google/gemma-3-4b-pt} \\
% \texttt{gemma-3-12b-pt}~\cite{team2025gemma} & \url{https://huggingface.co/google/gemma-3-12b-pt} \\
% \texttt{gemma-3-27b-pt}~\cite{team2025gemma} & \url{https://huggingface.co/google/gemma-3-27b-pt} \\
% \texttt{Llama-3.2-11B-Vision}~\cite{grattafiori2024llama} & \url{https://huggingface.co/meta-llama/Llama-3.2-11B-Vision} \\
% \texttt{Llama-3.2-90B-Vision}~\cite{grattafiori2024llama} & \url{https://huggingface.co/meta-llama/Llama-3.2-90B-Vision} \\
% \texttt{llava-1.5-7b-hf}~\cite{liu2024improved} & \url{https://huggingface.co/llava-hf/llava-1.5-7b-hf} \\
% \texttt{llava-1.5-13b-hf}~\cite{liu2024improved} & \url{https://huggingface.co/llava-hf/llava-1.5-13b-hf} \\
% \texttt{llava-v1.6-vicuna-7b-hf}~\cite{liu2024llavanext} & \url{https://huggingface.co/llava-hf/llava-v1.6-vicuna-7b-hf} \\
% \texttt{llava-v1.6-vicuna-13b-hf}~\cite{liu2024llavanext} & \url{https://huggingface.co/llava-hf/llava-v1.6-vicuna-13b-hf} \\
% \texttt{llava-v1.6-34b-hf} & \url{https://huggingface.co/llava-hf/llava-v1.6-34b-hf} \\
% \texttt{InternVL3-1B}~\cite{zhu2025internvl3} & \url{https://huggingface.co/OpenGVLab/InternVL3-1B} \\
% \texttt{InternVL3-8B}~\cite{zhu2025internvl3} & \url{https://huggingface.co/OpenGVLab/InternVL3-8B} \\
% \texttt{InternVL3-14B}~\cite{zhu2025internvl3} & \url{https://huggingface.co/OpenGVLab/InternVL3-14B} \\
% \bottomrule
% \label{tab:vlm_huggingface_link}
% \end{longtable}
% % }

\section{Implications of Visual Stitching on VLM Safety}\label{app:sec:moderation}

\begin{figure}[t]
  \centering
  \begin{subfigure}{\linewidth}
    \centering
    \includegraphics[trim=0 0 0 0, clip, width=\linewidth]{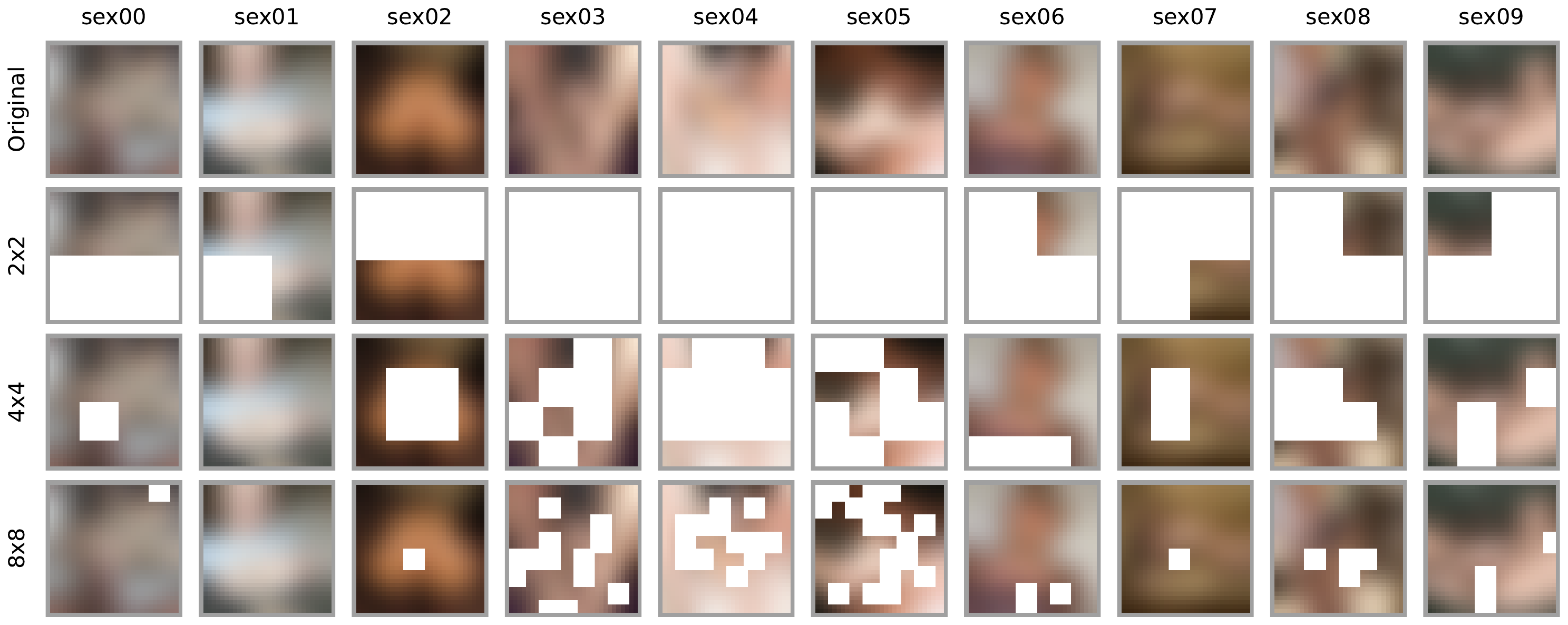}
    \caption{sex}
  \end{subfigure}

  \begin{subfigure}{\linewidth}
    \centering
    \includegraphics[trim=0 0 0 0, clip, width=\linewidth]{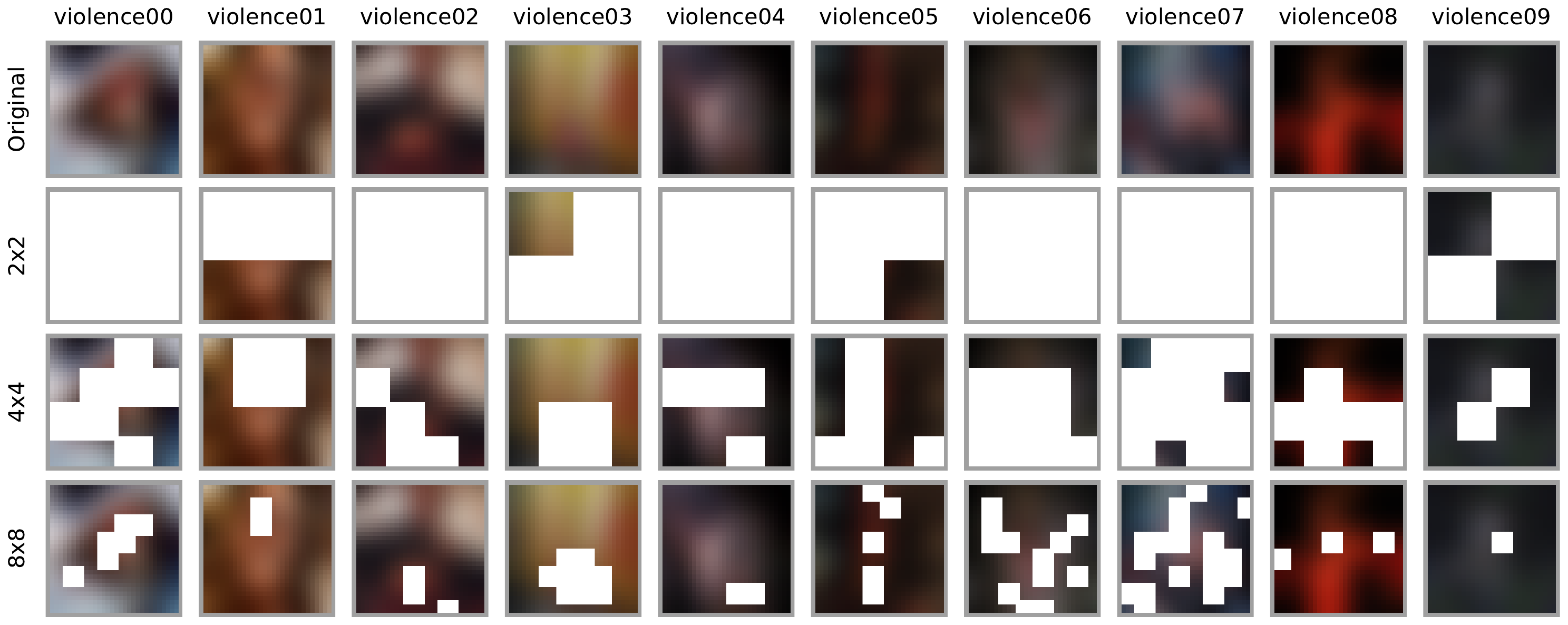}
    \caption{violence}
  \end{subfigure}
  \caption{Censored examples of $20$ dangerous images and their patches that evaded the OpenAI Moderation API (white patches indicate those flagged as dangerous).}
  \label{fig:moderation_visualization_all}
\end{figure}

\begin{figure}[t]
  \centering
  \includegraphics[trim=0 0 0 0, clip, width=\linewidth]{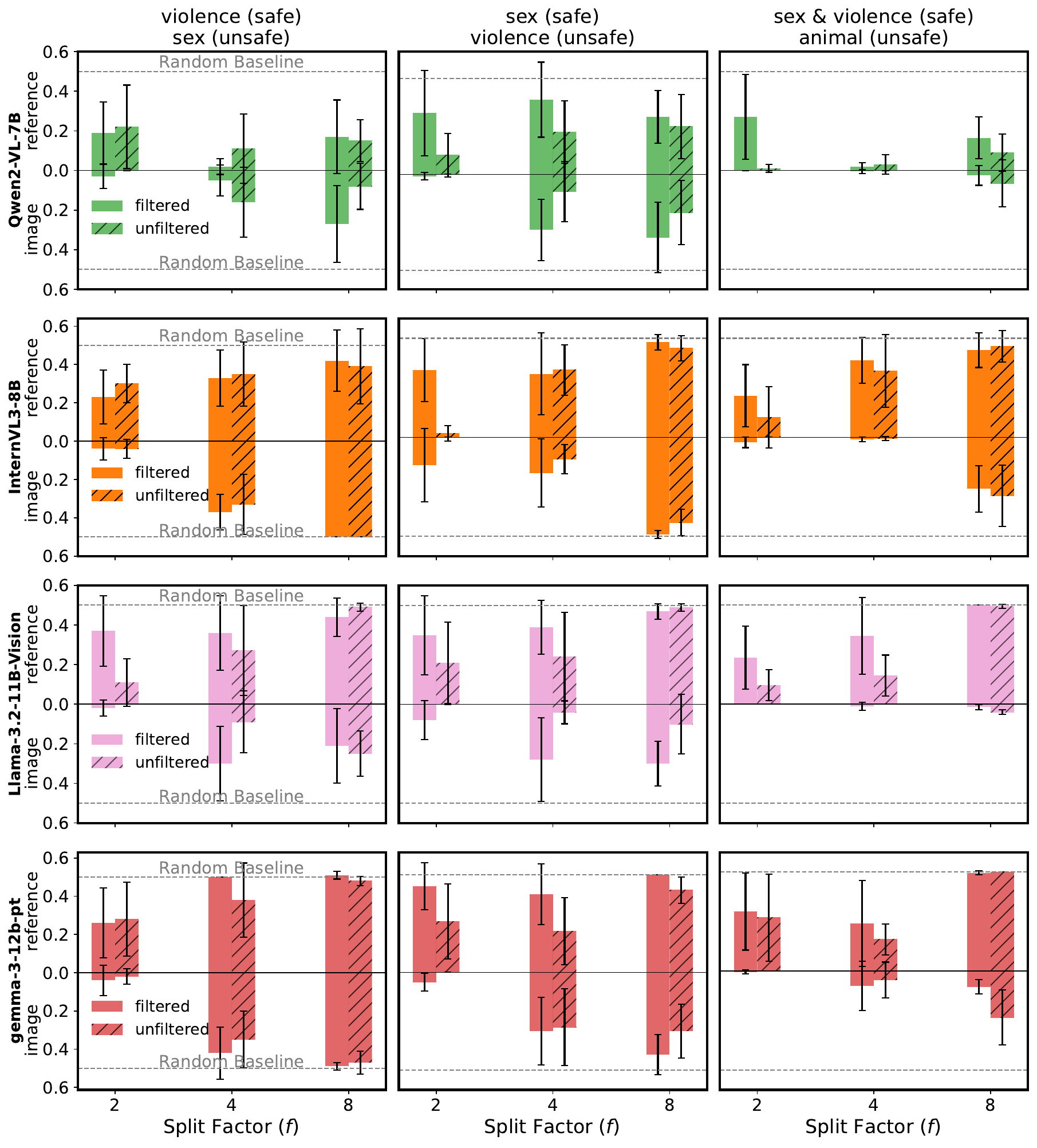} % Replace with your image file
  \caption{\textbf{Mean ranks of the correct \texttt{text} (lower is better) after finetuning different models on \texttt{(patch, text)} pairs, with and without OpenAI Moderation API filtering.} Lower ranks indicate successful emulation of direct tuning on the original \texttt{(image, text)} pairs, which would otherwise be flagged and discarded. See Figure~\ref{fig:moderation_all} for results on other models.}
  \label{fig:moderation_all}
\end{figure}

\subsection{Dataset Details}\label{app:subsec:moderation:dataset_details}
We construct a dataset of $20$ dangerous images: $10$ sex-related from the MultiTrust benchmark~\cite{zhang2024benchmarking}, and $10$ violence-related from horror films listed at \url{https://mikepwilliams-uk.tumblr.com/post/139723492184/10-of-the-goriest-deaths-in-horror-film-history}. Figure~\ref{fig:moderation_visualization_all} visualizes the censored version of these images as well as their patches that evade (i.e., classified as ``safe'') the OpenAI Moderation API~\cite{openai2024moderation}.

\subsection{Additional Results}\label{app:subsec:moderation:additional_results}

\paragraph{Finetuning on filtered patches enables harmful knowledge acquisition.} Figure~\ref{fig:moderation_qwen2_7b} in the main text presents results for \texttt{Qwen2-VL-7B}. Figure~\ref{fig:moderation_all} shows the full results for other models.

\clearpage
\iftoggle{isSubmission}{
  %%%%%%%%%%%%%%%%%%%%%%%%%%%%%%%%%%%%%%%%%%%%%%%%%%%%%%%%%%%%

\newpage
\section*{NeurIPS Paper Checklist}

%%% BEGIN INSTRUCTIONS %%%
The checklist is designed to encourage best practices for responsible machine learning research, addressing issues of reproducibility, transparency, research ethics, and societal impact. Do not remove the checklist: {\bf The papers not including the checklist will be desk rejected.} The checklist should follow the references and follow the (optional) supplemental material.  The checklist does NOT count towards the page
limit. 

Please read the checklist guidelines carefully for information on how to answer these questions. For each question in the checklist:
\begin{itemize}
    \item You should answer \answerYes{}, \answerNo{}, or \answerNA{}.
    \item \answerNA{} means either that the question is Not Applicable for that particular paper or the relevant information is Not Available.
    \item Please provide a short (1–2 sentence) justification right after your answer (even for NA). 
   % \item {\bf The papers not including the checklist will be desk rejected.}
\end{itemize}

{\bf The checklist answers are an integral part of your paper submission.} They are visible to the reviewers, area chairs, senior area chairs, and ethics reviewers. You will be asked to also include it (after eventual revisions) with the final version of your paper, and its final version will be published with the paper.

The reviewers of your paper will be asked to use the checklist as one of the factors in their evaluation. While "\answerYes{}" is generally preferable to "\answerNo{}", it is perfectly acceptable to answer "\answerNo{}" provided a proper justification is given (e.g., "error bars are not reported because it would be too computationally expensive" or "we were unable to find the license for the dataset we used"). In general, answering "\answerNo{}" or "\answerNA{}" is not grounds for rejection. While the questions are phrased in a binary way, we acknowledge that the true answer is often more nuanced, so please just use your best judgment and write a justification to elaborate. All supporting evidence can appear either in the main paper or the supplemental material, provided in appendix. If you answer \answerYes{} to a question, in the justification please point to the section(s) where related material for the question can be found.

IMPORTANT, please:
\begin{itemize}
    \item {\bf Delete this instruction block, but keep the section heading ``NeurIPS Paper Checklist"},
    \item  {\bf Keep the checklist subsection headings, questions/answers and guidelines below.}
    \item {\bf Do not modify the questions and only use the provided macros for your answers}.
\end{itemize}

%%% END INSTRUCTIONS %%%

\begin{enumerate}

\item {\bf Claims}
    \item[] Question: Do the main claims made in the abstract and introduction accurately reflect the paper's contributions and scope?
    \item[] Answer: \answerYes{} % Replace by \answerYes{}, \answerNo{}, or \answerNA{}.
    \item[] Justification: See Section~\ref{sec:experiments} and Section~\ref{sec:moderation}.

\item {\bf Limitations}
    \item[] Question: Does the paper discuss the limitations of the work performed by the authors?
    \item[] Answer: \answerYes{} % Replace by \answerYes{}, \answerNo{}, or \answerNA{}.
    \item[] Justification: See Section~\ref{sec:discussion_and_limitations}.

\item {\bf Theory assumptions and proofs}
    \item[] Question: For each theoretical result, does the paper provide the full set of assumptions and a complete (and correct) proof?
    \item[] Answer: \answerNA{} % Replace by \answerYes{}, \answerNo{}, or \answerNA{}.
    \item[] Justification: No theoretical result is involved.

    \item {\bf Experimental result reproducibility}
    \item[] Question: Does the paper fully disclose all the information needed to reproduce the main experimental results of the paper to the extent that it affects the main claims and/or conclusions of the paper (regardless of whether the code and data are provided or not)?
    \item[] Answer: \answerYes{} % Replace by \answerYes{}, \answerNo{}, or \answerNA{}.
    \item[] Justification: See Section~\ref{sec:experiments}, Section~\ref{sec:moderation}, Appendix~\ref{app:sec:experiments} and Appendix~\ref{app:sec:moderation}.

\item {\bf Open access to data and code}
    \item[] Question: Does the paper provide open access to the data and code, with sufficient instructions to faithfully reproduce the main experimental results, as described in supplemental material?
    \item[] Answer: \answerYes{} % Replace by \answerYes{}, \answerNo{}, or \answerNA{}.
    \item[] Justification: The data and code will be included in the supplemental material.

\item {\bf Experimental setting/details}
    \item[] Question: Does the paper specify all the training and test details (e.g., data splits, hyperparameters, how they were chosen, type of optimizer, etc.) necessary to understand the results?
    \item[] Answer: \answerYes{} % Replace by \answerYes{}, \answerNo{}, or \answerNA{}.
    \item[] Justification: See Appendix~\ref{app:subsec:experiments:training_details}.

\item {\bf Experiment statistical significance}
    \item[] Question: Does the paper report error bars suitably and correctly defined or other appropriate information about the statistical significance of the experiments?
    \item[] Answer: \answerYes{} % Replace by \answerYes{}, \answerNo{}, or \answerNA{}.
    \item[] Justification: We include error bars in our figures.

\item {\bf Experiments compute resources}
    \item[] Question: For each experiment, does the paper provide sufficient information on the computer resources (type of compute workers, memory, time of execution) needed to reproduce the experiments?
    \item[] Answer: \answerYes{} % Replace by \answerYes{}, \answerNo{}, or \answerNA{}.
    \item[] Justification: See Appendix~\ref{app:subsec:experiments:training_details}.
    
\item {\bf Code of ethics}
    \item[] Question: Does the research conducted in the paper conform, in every respect, with the NeurIPS Code of Ethics \url{https://neurips.cc/public/EthicsGuidelines}?
    \item[] Answer: \answerYes{} % Replace by \answerYes{}, \answerNo{}, or \answerNA{}.
    \item[] Justification: We have reviewed NeurIPS Code of Ethics and made sure our research conforms with it.

\item {\bf Broader impacts}
    \item[] Question: Does the paper discuss both potential positive societal impacts and negative societal impacts of the work performed?
    \item[] Answer: \answerYes{} % Replace by \answerYes{}, \answerNo{}, or \answerNA{}.
    \item[] Justification: See Abstract, Section~\ref{sec:introduction} and Section~\ref{sec:conclusion}
    
\item {\bf Safeguards}
    \item[] Question: Does the paper describe safeguards that have been put in place for responsible release of data or models that have a high risk for misuse (e.g., pretrained language models, image generators, or scraped datasets)?
    \item[] Answer: \answerYes{} % Replace by \answerYes{}, \answerNo{}, or \answerNA{}.
    \item[] Justification: See Section~\ref{sec:introduction} and Section~\ref{sec:conclusion}.

\item {\bf Licenses for existing assets}
    \item[] Question: Are the creators or original owners of assets (e.g., code, data, models), used in the paper, properly credited and are the license and terms of use explicitly mentioned and properly respected?
    \item[] Answer: \answerYes{} % Replace by \answerYes{}, \answerNo{}, or \answerNA{}.
    \item[] Justification: We have carefully cited all papers that our work builds upon or is closely related to.

\item {\bf New assets}
    \item[] Question: Are new assets introduced in the paper well documented and is the documentation provided alongside the assets?
    \item[] Answer: \answerYes{} % Replace by \answerYes{}, \answerNo{}, or \answerNA{}.
    \item[] Justification: The data and code in the supplementary material have been well documented. 

\item {\bf Crowdsourcing and research with human subjects}
    \item[] Question: For crowdsourcing experiments and research with human subjects, does the paper include the full text of instructions given to participants and screenshots, if applicable, as well as details about compensation (if any)? 
    \item[] Answer: \answerNA{} % Replace by \answerYes{}, \answerNo{}, or \answerNA{}.
    \item[] Justification: The paper does not involve crowdsourcing nor research with human subjects.

\item {\bf Institutional review board (IRB) approvals or equivalent for research with human subjects}
    \item[] Question: Does the paper describe potential risks incurred by study participants, whether such risks were disclosed to the subjects, and whether Institutional Review Board (IRB) approvals (or an equivalent approval/review based on the requirements of your country or institution) were obtained?
    \item[] Answer: \answerNA{} % Replace by \answerYes{}, \answerNo{}, or \answerNA{}.
    \item[] Justification: The paper does not involve crowdsourcing nor research with human subjects.

\item {\bf Declaration of LLM usage}
    \item[] Question: Does the paper describe the usage of LLMs if it is an important, original, or non-standard component of the core methods in this research? Note that if the LLM is used only for writing, editing, or formatting purposes and does not impact the core methodology, scientific rigorousness, or originality of the research, declaration is not required.
    %this research? 
    \item[] Answer: \answerNA{} % Replace by \answerYes{}, \answerNo{}, or \answerNA{}.
    \item[] Justification: The core method development in this research does not involve LLMs as any important, original, or non-standard components.

\end{enumerate}

}

\end{document}